\documentclass{article}

\usepackage{PRIMEarxiv}

\usepackage[utf8]{inputenc} %
\usepackage[T1]{fontenc}    %
\usepackage{hyperref}       %
\usepackage{url}            %
\usepackage{booktabs}       %
\usepackage{amsfonts}       %
\usepackage{amsmath}       %
\usepackage{nicefrac}       %
\usepackage{microtype}      %
\usepackage{lipsum}
\usepackage{fancyhdr}       %
\usepackage{graphicx}       %
\graphicspath{{media/}}     %
\usepackage{natbib}
\usepackage{todonotes}
\usepackage{cleveref}

\usepackage{amsthm}
\usepackage{caption,listings}
\usepackage{tcolorbox}

\theoremstyle{definition}
\newtheorem{definition}{Definition}[section]

\newcommand{\myexample}[2]{
    \begin{tcolorbox}[colback=black!5!white,colframe=black,title={#1}]
        #2
    \end{tcolorbox}
}

\usepackage{wrapfig}

\usepackage{subfig}
\def\X{\mathbf{X}}

\newcommand{\eg}{\textit{e.g\@.}}

\newcommand{\ie}{\textit{i.e\@.}}
\newcommand{\md}{\textit{model collapse}}
\newcommand{\Md}{\textit{Model Collapse}}

\usepackage[]{mdframed}

\usepackage{tcolorbox}

\title{The Curse of Recursion: \\Training on Generated Data Makes Models Forget}

\author{
  Ilia Shumailov*\\
  University of Oxford\\
  \And
  Zakhar Shumaylov* \\
  University of Cambridge\\
  \And
  Yiren Zhao \\
  Imperial College London\\
  \And 
  Yarin Gal \\
  University of Oxford \\
  \And 
  Nicolas Papernot \\
  University of Toronto \& Vector Institute\\
  \And 
  Ross Anderson \\
  University of Cambridge \& University of Edinburgh\\
}

\begin{document}
\maketitle

\begin{abstract}
Stable Diffusion revolutionised image creation from descriptive text. \texttt{GPT-2}, \texttt{GPT-3(.5)} and \texttt{GPT-4} demonstrated astonishing performance across a variety of language tasks. \texttt{ChatGPT} introduced such language models to the general public. It is now clear that large language models (LLMs) are here to stay, and will bring about drastic change in the whole ecosystem of online text and images. In this paper we consider what the future might hold. What will happen to \texttt{GPT-$\{n\}$} once LLMs contribute much of the language found online? We find that use of model-generated content in training causes irreversible defects in the resulting models, where tails of the original content distribution disappear. We refer to this effect as~\md\footnote{The name is inspired by the Generative Adversarial Networks (GAN) literature on mode collapse, where GANs start producing a limited set of outputs that all trick the discriminator. \Md~is a process whereby models eventually converge to a state similar to that of a GAN Mode Collapse. The original version of this paper referred to this effect as `model dementia', but we decided to change this following feedback that it trivialised the medical notion of `dementia' and could cause offence. }~and show that it can occur in Variational Autoencoders, Gaussian Mixture Models and LLMs. We build theoretical intuition behind the phenomenon and portray its ubiquity amongst all learned generative models. We demonstrate that it has to be taken seriously if we are to sustain the benefits of training from large-scale data scraped from the web. Indeed, the value of data collected about genuine human interactions with systems will be increasingly valuable in the presence of content generated by LLMs in data crawled from the Internet.
\end{abstract}

\section{Introduction}

A lot of human communication happens online. Billions of emails are exchanged daily, along with billions of social-media messages and millions of news articles. Almost all of this material was produced and curated only by humans in the early years of the worldwide web, yet since the turn of the century search engines have come to determine what people can find, and in the past decade smart text editors with spelling and grammar correction have helped tweak what we produce. %
Now, text can not only be groomed and analysed efficiently; it can also be generated -- by large language models (LLMs). These models now (arguably) pass a weaker form of the Turing test in the sense that their output cannot be reliably distinguished from text written by humans~\citep{solaiman2019release}.

The development of LLMs is quite involved and requires masses of training data. Anecdotally, some powerful recent models are trained using scrapes of much of the Internet, then further fine-tuned with reinforcement learning from human feedback (RLHF)~\citep{NIPS2013_e034fb6b,openai2023gpt4}. This further boosts the effective dataset size. Yet while current LLMs \citep{devlin2018bert, liu2019roberta, brown2020language, zhang2022opt}, including \texttt{GPT-4}, were trained on predominantly human-generated text, this may change in the future. If most future models' training data is also scraped from the web, then they will inevitably come to train on data produced by their predecessors. In this paper, we investigate what happens when text produced, \eg~by a version of \texttt{GPT}, forms most of the training dataset of following models. What happens to \texttt{GPT} versions \texttt{GPT-\{$n$\}} as generation $n$ increases?\footnote{This is not limited to text models; one can also consider what happens when music created by human composers and played by human musicians trains models whose output trains other models.}

We discover that learning from data produced by other models causes \md~-- a degenerative process whereby, over time, models forget the true underlying data distribution, even in the absence of a shift in the distribution over time. We give examples of \md~for Gaussian Mixture Models (GMMs), Variational Autoencoders (VAE) and Large Language models (LLMs). We show that over time we start losing information about the true distribution, which first starts with tails disappearing, and over the generations learned behaviours start converging to a point estimate with very small variance. Furthermore, we show that this process is inevitable, even for cases with almost ideal conditions for long-term learning \ie~no function estimation error.

\begin{wrapfigure}{r}{0.5\linewidth}
  \begin{center}
    \includegraphics[width=\linewidth]{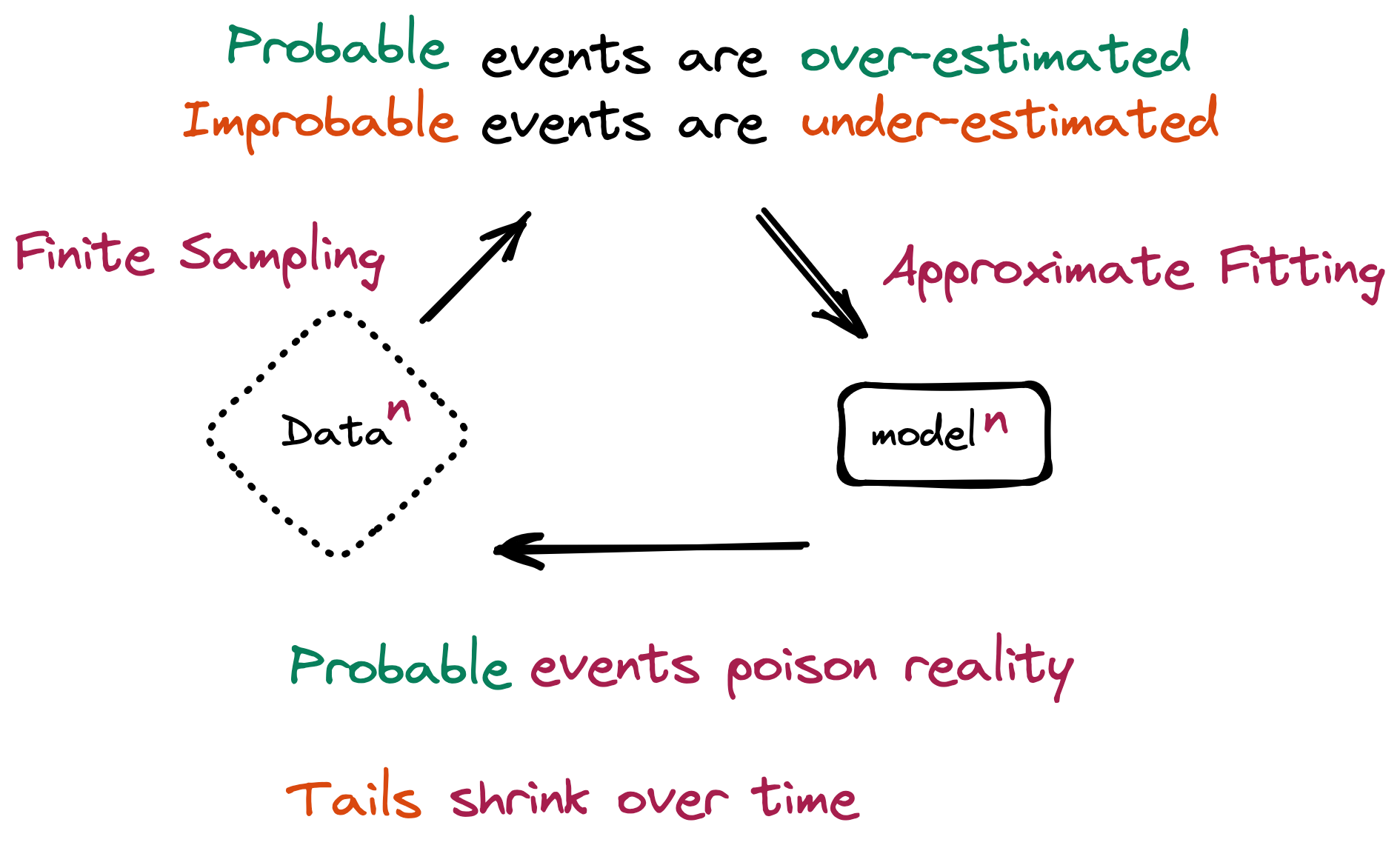}
  \end{center}
  \caption{\Md~refers to a degenerative learning process where models start forgetting improbable events over time, as the model becomes poisoned with its own projection of reality.}
\end{wrapfigure}

Finally, we discuss the broader implications of~\md. We note that access to the original data distribution is crucial: in learning where the tails of the underlying distribution matter, one needs access to real human-produced data. In other words, the use of LLMs at scale to publish content on the Internet will pollute the collection of data to train them: data about human interactions with LLMs will be increasingly valuable.%

This paper is structured as follows. First, in~\Cref{sec:background,sec:theoreticalint} we describe the reasons why \md~happens. To best describe the intuition, we present a simple example of a single-dimensional Gaussian where errors due to sampling inevitably cause~\md, which are then extended to a multidimensional generative model under some assumptions. Under both models, similar lower bounds are derived on the risk, defined in terms of the Wasserstein distance from the true distribution. Next, we turn to GMMs and VAEs to show that additional functional approximation errors further exacerbate~\md. Finally, we discuss the most commonly used setting of fine-tuned language models, where we report that only early signs of~\md~can be detected, if models are fine-tuned as opposed to trained from scratch.   

In this paper we make the following contributions:

\begin{itemize}
    \item We demonstrate the existence of a degenerative process in learning and name it \md; 
    \item We demonstrate that \md~exists in a variety of different model types and datasets;
    \item We show that, to avoid \md, access to genuine human-generated content is essential.
\end{itemize}

\section{Related work}
In this section we are going to cover two closest concepts to \md~from existing literature: catastrophic forgetting and data poisoning. Neither is able to explain the phenomenon of \md~fully, as the setting is fundamentally different, but they provide another perspective on the observed phenomenon. 
\subsection{Continual learning and catastrophic forgetting}
Unlike traditional machine learning which seeks to learn from a static data distribution, \emph{continual learning} attempts to learn from a dynamic one, where data are supplied in a sequential fashion \citep{van2019three}. This tends to be task-based, where data are provided with delineated task boundaries; \eg, classifying dogs from cats and recognising handwritten digits. Our work is more similar to task-free continual learning \citep{aljundi2019task} where data distributions gradually change without the notion of separate tasks. Our work examines a particular scenario in which the changed data distributions arise from the model itself, as a result of training in the previous iteration.

\myexample{Example of text outputs of an OPT-125m model affected by~\Md -- models degrade over generations, where each new generation is trained on data produced by the previous generation.}{
\begin{description}
    \item \textbf{Input:} some started before 1360 — was typically accomplished by a master mason and a small team of itinerant masons, supplemented by local parish labourers, according to Poyntz Wright. But other authors reject this model, suggesting instead that leading architects designed the parish church towers based on early examples of Perpendicular
    \item \textbf{Outputs:}
    
    \item \textbf{~Gen 0:} Revival architecture such as St. John's Cathedral in London. The earliest surviving example of Perpendicular Revival architecture is found in the 18th @-@ century Church of Our Lady of Guernsey, which dates from the late 19th century. There are two types of perpendicular churches : those

    \item \textbf{~Gen 1:}  architecture such as St. Peter's Basilica in Rome or St. Peter's Basilica in Buenos Aires. There is no evidence that any of these buildings were built during the reign of Pope Innocent III, but it is possible that they may have been built during the reign of his successor, Pope Innocent

    \item \textbf{~Gen 7:}  architecture in England. In an interview with The New York Times, Wright said : " I don 't think there is anything wrong with me being able to do what I want to do. It just doesn 't work for me. " He added : " I don 't know if you can call it funny,

    \item \textbf{~Gen 9:} architecture. In addition to being home to some of the world's largest populations of black @-@ tailed jackrabbits, white @-@ tailed jackrabbits, blue @-@ tailed jackrabbits, red @-@ tailed jackrabbits, yellow @-
\end{description}
}

A typical challenge in continual learning is that the model forgets previous samples when learning new information; this is known as \emph{catastrophic forgetting} \citep{kirkpatrick2017overcoming}. A typical way of preventing it is to use regularisations (Memory Aware Synpass \citep{aljundi2018memory}) or just rely on data (\eg~Learning without Forgetting \citep{li2017learning}). This has an indirect connection to our work, yet differs since the data in the process of \md~are generated by different generations of models.

\subsection{Data poisoning}
Poisoning attacks are crafted and inserted during training in order to degrade the model's performance when deployed~\citep{biggio2012poisoning}. Malicious data can be inserted into training data to induce unintended behaviors that can be activated by special triggers~\citep{gu2017badnets}. The early literature on data poisoning focused mainly on supervised learning, where classifiers are trained with labeled samples. But with the emergence of contrastive learning~\citep{radford2021learning} and LLMs~\citep{brown2020language}, more recent models are trained with large-scale web crawls, making data poisoning attacks more feasible on these untrustworthy web sources. Recent studies have demonstrated that web-scale datasets can be poisoned by introducing malicious data into a small percentage of samples~\citep{carlini2021poisoning, carlini2023poisoning}.

\section{What is \Md?}
\label{sec:background}

\begin{figure}[t]
\centering
\includegraphics[width=.8\linewidth]{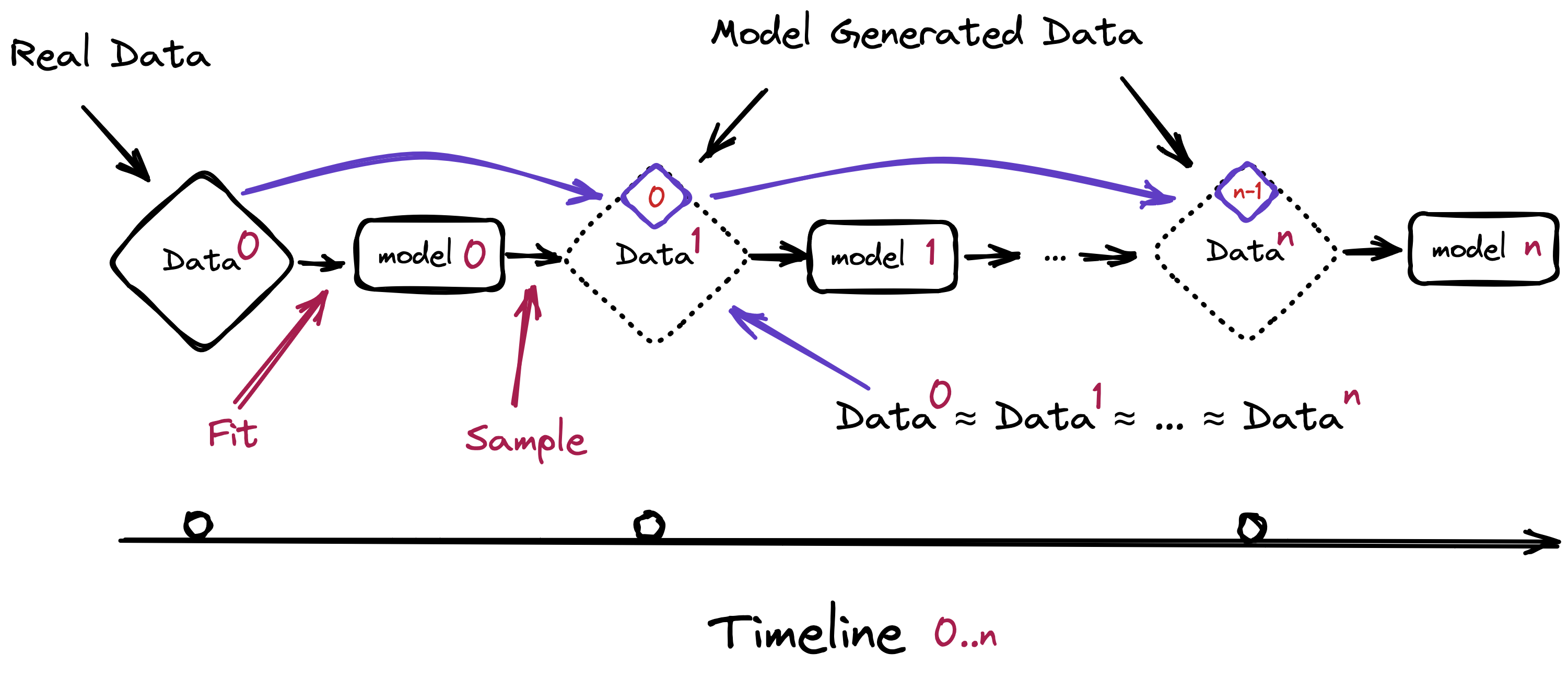}
\caption{The high-level description of the feedback mechanism in the learning process. Here, data are assumed to be human-curated and start off clean; then model $0$ is trained and data are sampled from it; at step $n$, data are added to the overall data from step $n-1$, and this ensemble is used to train model $n$.  Data obtained with Monte Carlo sampling should ideally be statistically close to the original, provided \textit{fitting} and \textit{sampling} procedures are perfect. This process depicts what happens in real life with the Internet -- model-generated data become pervasive. }
\label{fig:highlevel_disc}
\end{figure}

\begin{definition}[\Md]
\Md~is a degenerative process affecting generations of learned generative models, where generated data end up polluting the training set of the next generation of models; being trained on polluted data, they then mis-perceive reality. We separate two special cases: \textbf{early~\md} and \textbf{late~\md}. In early~\md~the model begins losing information about the tails of the distribution; in the late~\md~model entangles different modes of the original distributions and converges to a distribution that carries little resemblance to the original one, often with very small variance. 
\end{definition}

Note that this process is different from the process of \textit{catastrophic forgetting} in that we are considering multiple models over time, in which our models do not forget previously learned data, but rather start misinterpreting what they believe to be real, by reinforcing their own beliefs.

This process occurs due to two specific sources of error compounding over generations and causing deviation from the original model. Of these, one source of error plays a primary role, and in the absence of it, the process would not occur beyond the first generation.

\subsection{Causes of \md}
\label{sec:sourceerror}

There are two main causes for \md, one primary and one secondary, which we describe now. Further mathematical intuition is provided in \Cref{sec:theoreticalint} to explain how these give rise to the errors observed, how different sources can compound and how we can quantify the average model divergence rate. 
\begin{itemize}
    \item \textit{Statistical approximation error} -- this is the primary type of error, which arises due to the number of samples being finite, and disappears as the number of samples tends to infinity. This occurs due to a non-zero probability that information can get lost at every step of re-sampling. \Cref{fig:single_dim_gaus_approx} shows an example of an approximation error. Here, a single-dimensional Gaussian is being approximated from a finite number of samples. Despite using a very large number of points, the errors remain significant; with $10^7$ samples we estimate the mean to be $0.00024899 \pm 1.89382984e^{-4}$, when the true value is $0$.
    \item \textit{Functional approximation error} -- this is a secondary type of error, which stems from our function approximators being insufficiently expressive (or sometimes too expressive outside of the original distribution support~\citep{nguyen2015deep}). It is well known that neural networks are universal functional approximators in the limit, but in practice this is not always true. In particular, a neural network can introduce non-zero likelihood outside of the support of the original distribution. A simple example of this error is if we were to try fitting a mixture of two Gaussians with a single Gaussian. Even if we have perfect information about the data distribution, model errors will be inevitable. It is important to also note that in the absence of statistical error, functional approximation error only occurs at the first generation. Once the new distribution belongs to the image of functional approximator, it remains exactly the same over the generations.
\end{itemize}

Each of the above can cause \md~to get worse or better. Better approximation power can even be a double-edged sword -- better expressiveness may counteract statistical noise, resulting in a good approximation of the true distribution, but it can equally compound this noise. More often then not, we get a cascading effect where combined individual inaccuracy causes the overall error to grow. Overfitting the density model will cause the model to extrapolate incorrectly and might give high density to low-density regions not covered in the training set support; these will then be sampled with arbitrary frequency.

It is worth mentioning that modern computers also have a further computational error coming from the way floating point numbers are represented. This error is not evenly spread across different floating point ranges, making it hard to estimate the precise value of a given number. Such errors are smaller in magnitude and are fixable with more precise hardware, making them less influential on~\md.

\section{Theoretical intuition}
\label{sec:theoreticalint}

In this section we aim to provide a theoretical intuition for the phenomenon of \md. We argue that the process of \md~is universal among generative models that recursively train on data generated by previous generations. We construct toy mathematical models, which prove to be simple enough to provide analytical expressions for quantities of interest, but also portray the phenomenon of \md. We aim to quantify how different sources of error can affect the overall end approximation of the original distribution. As discussed in \Cref{sec:sourceerror}, there are two main sources we are interested in -- \textit{statistical} error and \textit{functional} error. Since in the real world one rarely has infinite samples, quantifying the functional approximation error alone is of little interest for discussion of \md. Therefore, we will examine two simple cases: a discrete distribution in the absence of functional approximation error and a single dimensional Gaussian case, which portrays how functional approximation error can compound with statistical error. 

The overall stochastic process we are going to be considering (which we call \textit{Learning with Generational Data}) is the following. Assume that at generation $i$ we have a dataset $\mathcal{D}_i$ comprising of i.i.d. random variables $X^i_j$, where $j \in \{1,\dots,M_i\}$ denotes the sample number at generation $i$ and $M_i\geq2$. We will denote the distribution of $X^i$ as $p_i$. Here we assume that $p_0$ denotes the original distribution, from which the data comes from. Going from generation $i$ to generation $i+1$, we aim to estimate the distribution of samples in $\mathcal{D}_i$, with an approximation $p_{\theta_{i+1}}$. This step is what we refer to as functional approximation $\mathcal{F_\theta}: p_i \to p_{\theta_{i+1}}$. We then resample the dataset $\mathcal{D}_{i+1}$ from the distribution $p_{i+1} = \alpha_i p_{\theta_{i+1}} + \beta_i p_{i} + \gamma_ip_0$, with non-negative parameters $\alpha_i, \beta_i, \gamma_i$ summing up to $1$, \ie~they represent proportions of data used from different generations. This corresponds to a mixing of data coming from the original distribution ($\gamma_i$), data used by the previous generation ($\beta_i$) and data generated by the new model ($\alpha_i$). We refer to this as the sampling step. For the mathematical models to come, we consider $\alpha_i=\gamma_i=0$ \ie~data only from a single step is used, while numerical experiments are performed on more realistic choices of parameters. 

\subsection{Discrete distributions with exact approximation}
\begin{figure}
    \centering
    \includegraphics[width=0.9\linewidth]{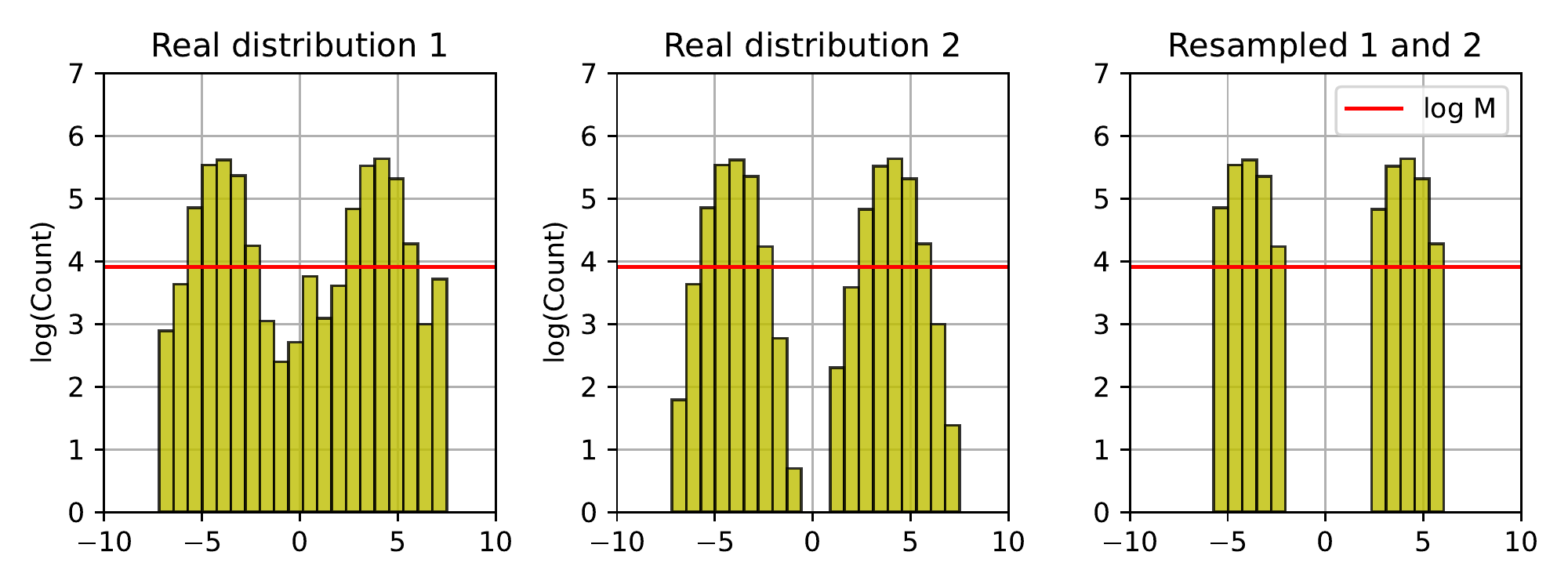}
    \caption{Shown in the middle is a histogram plot of samples from a Gaussian mixture with means $(-4,4)$ and variances of $1$. To the left of it is a similar distribution, but with 'fatter' tails, and on the right the same histograms are shown, but with low probability events being cut off due to finite resampling. Although distributions 1 and 2 are very different, when resampled (only assuming the expected behaviour), the tails get cut off, leading to the same observed distribution. In this case this is all states with probability less than $1/M$, or equivalently, bins with $\log{\texttt{Count}}\leq\log{M}$.}
    \label{fig:histograms}
\end{figure}

In this subsection we consider a discrete probability distribution, which is represented by a histogram, \eg~as shown on \Cref{fig:histograms}. In what follows we consider the stochastic process in absence of functional approximation error, \ie~\;$\mathcal{F}(p)=p$. In this case, \md~arises only due to statistical errors from the sampling step. At first, the tails (low probability events) begin to disappear due to low probability of sampling them, and over time the distribution becomes a delta function. Denoting the sample size as $M$, if we consider state $i$ with probability $q\leq\frac{1}{M}$, the expected number of samples with value $i$ coming from those events will be less than $1$, which means that in practice we will lose information about them. This is portrayed on \Cref{fig:histograms}, where infrequent events get cut off. Considering more generally some state $i$ with probability $q$, using standard conditional probability one can show that the probability of losing information (\ie~sampling no data at some generation) is equal to $1-q$. But this in turn means that we must converge to a delta function positioned at some state, and the probability of ending up at a certain state is equal to the probability of sampling said state from the original distribution.

But how do we show directly that this process is going to turn our distribution into a delta function? By considering the process as going from $\mathbf{X}^i\to\mathcal{F_\theta}\to p_{i+1}\to\mathbf{X}^{i+1}$, we see that this forms a Markov Chain, as $\mathbf{X}^{i+1}$ only depends on $\mathbf{X}^{i}$. Furthermore, if all the $X^{i}_j$ have the same value, then at the next generation the approximated distribution will be exactly a delta function, and therefore all of $X^{i+1}_j$ will also have the same value. 
This implies that the Markov chain contains at least one absorbing state, and therefore with probability 1 it will converge to one of the absorbing states. This is a well-known fact, of which a proof is provided in \Cref{ap:markov}. For this chain, the only absorbing states are those corresponding to delta functions. As a result, as we follow the progress of \md, we are guaranteed  to end up in a constant state, having lost all the information of the original distribution when the chain is absorbed.\footnote{This argument also works in general due to floating point representations being discrete, making the Markov Chain over the parameters of the model discrete. Thus as long as the model parameterisation allows for delta functions, we \textit{will} get to it, as due to sampling errors the only possible absorbing states are delta functions.} Based on the discussion above we see how both early and late stage \md~ must arise in the case of discrete distributions with perfect functional approximation.

\begin{figure}%
    \centering
    \subfloat[\centering Mean estimation]{{\includegraphics[width=0.43\textwidth]{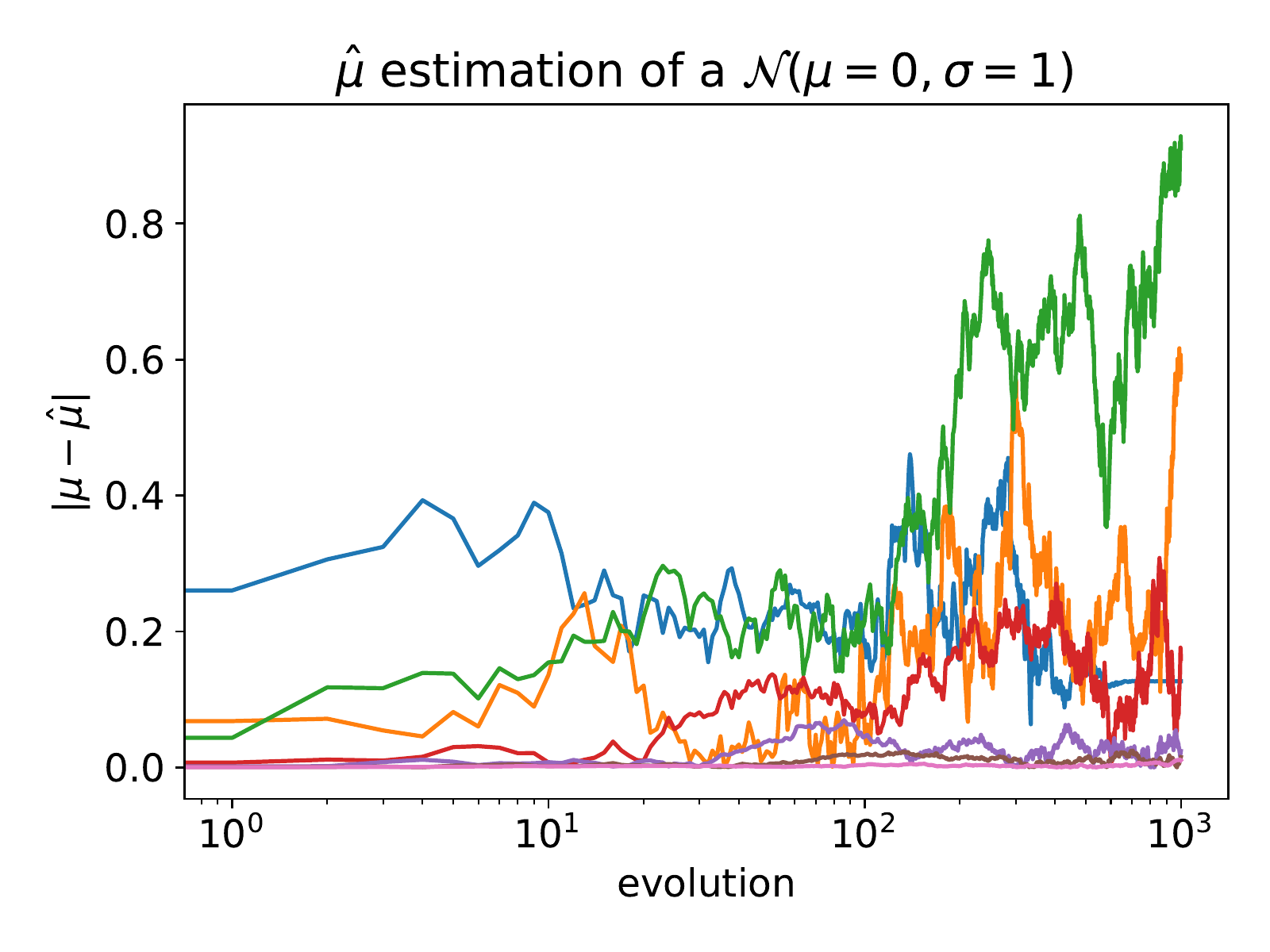} }}%
    \subfloat[\centering Standard Deviation]{{\includegraphics[width=0.43\textwidth]{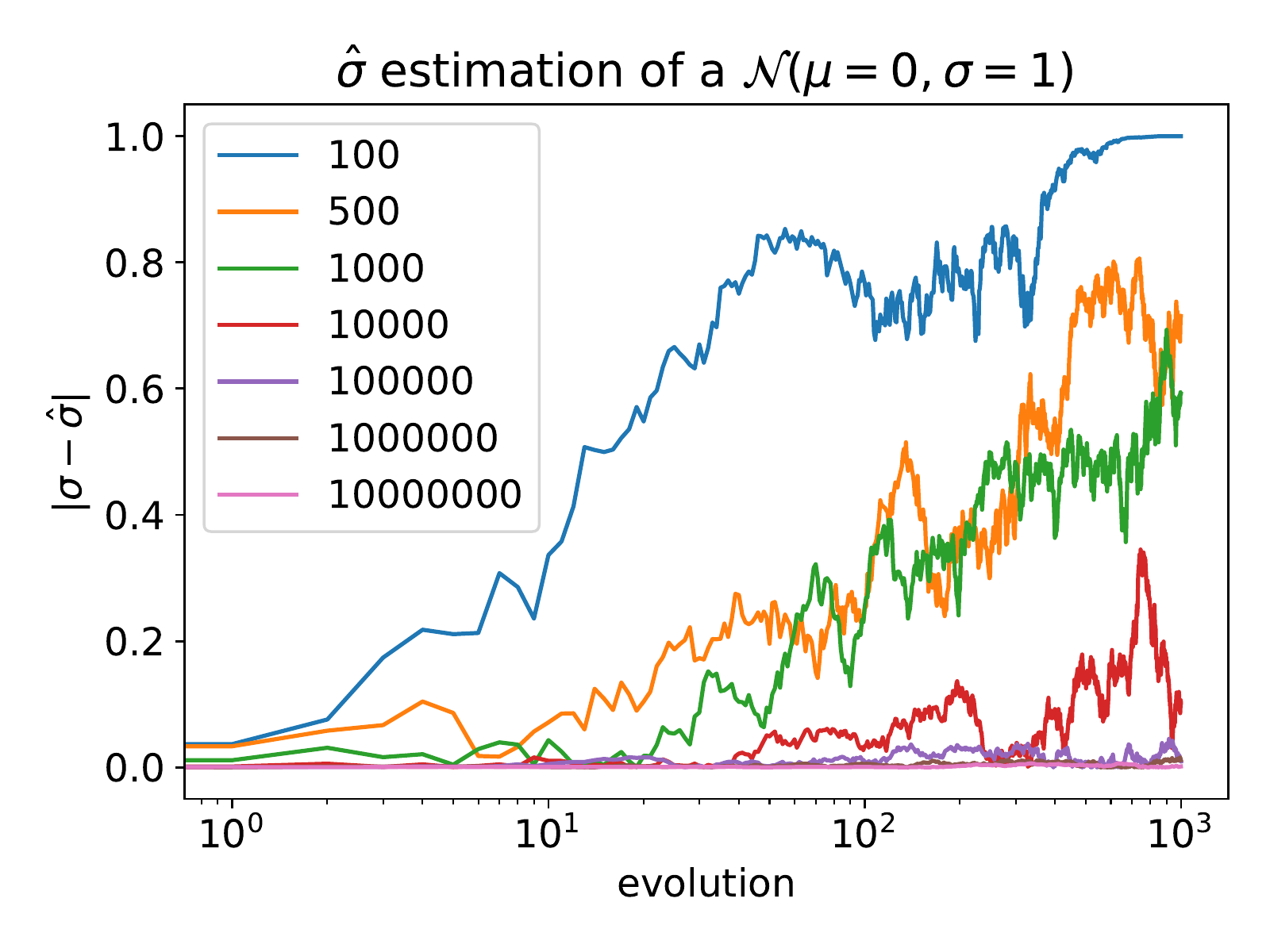} }}%
    \caption{Recursive fitting-sampling of a 1D Gaussian with different numbers of samples drawn. We find that unless sampled a very large number of times,~\ie~<100000, both standard deviation and mean get significantly affected. Here we report a single run; while re-running the experiment changes the initial performance, both $\mu$ and $\sigma$ drift over time. The overall graph looks quite similar to that of a Gaussian random walk.}%
    \label{fig:example1}%

    \centering
    \subfloat[\centering Mean estimation]{{\includegraphics[width=0.43\textwidth]{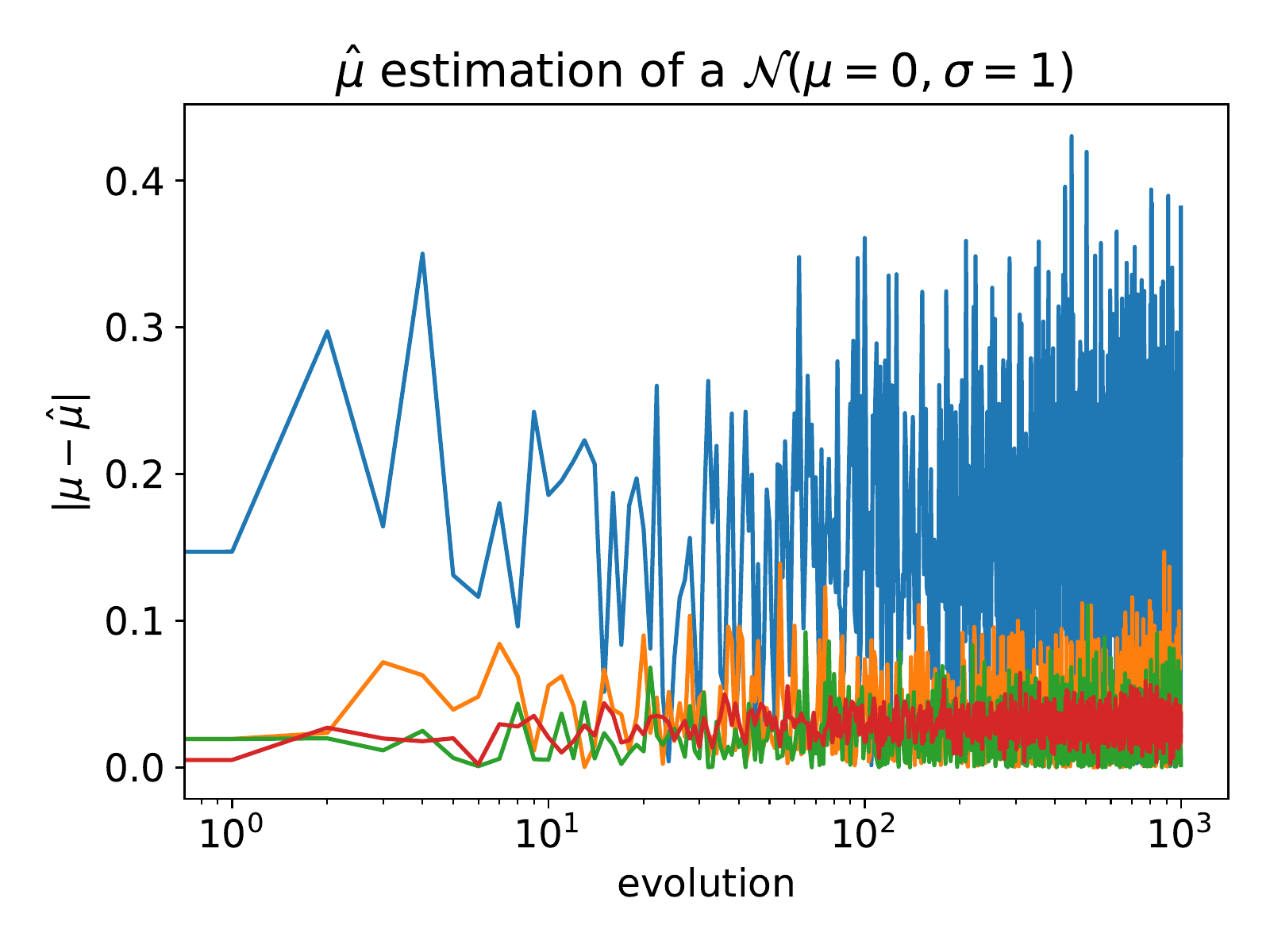} }}%
    \subfloat[\centering Standard Deviation]{{\includegraphics[width=0.43\textwidth]{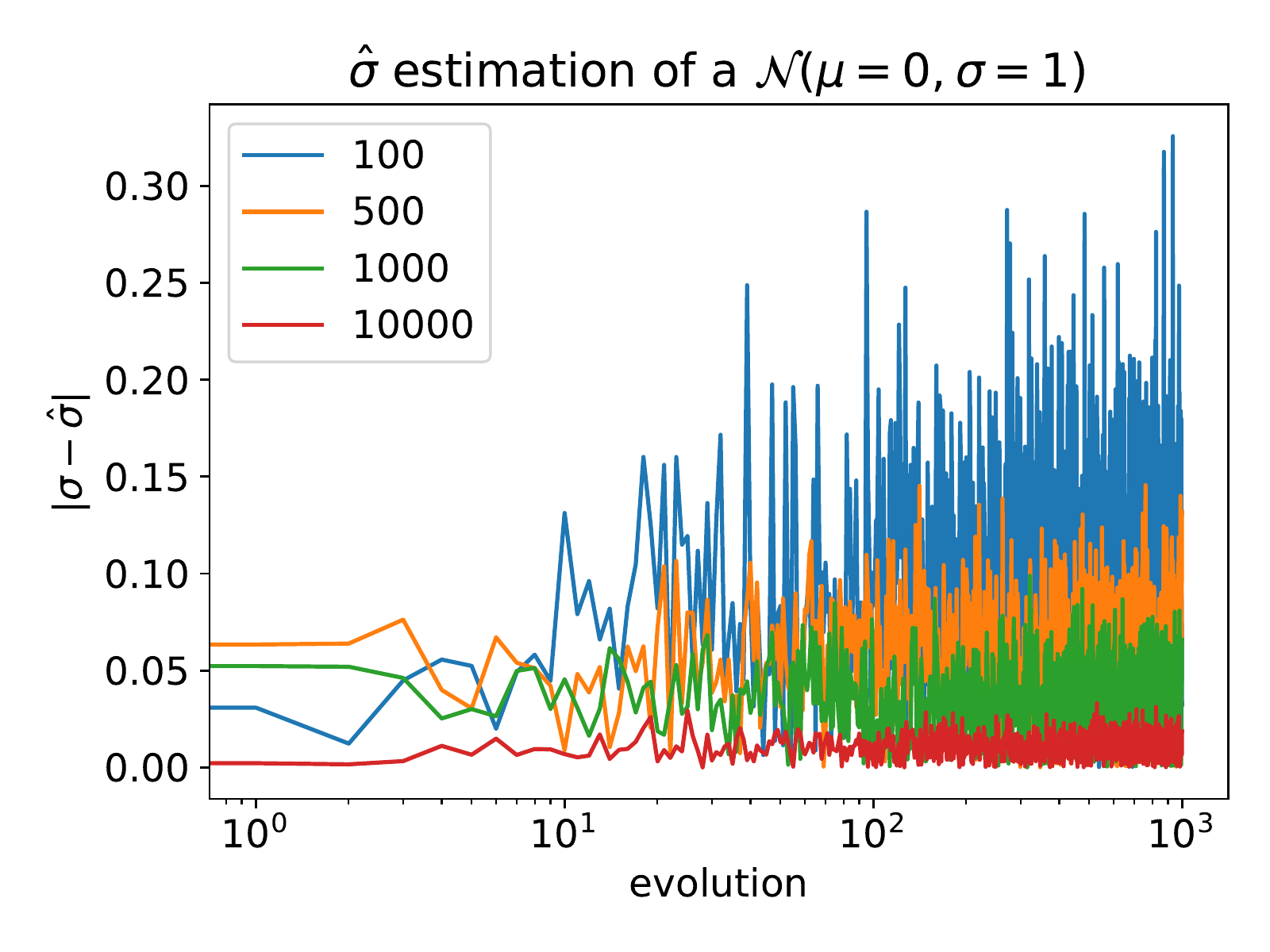} }}%
    \caption{Recursive fitting-sampling of a 1D Gaussian with different numbers of samples drawn. In this plot data get accumulated in a pool, from which a fixed sample is drawn. In other words, a model $n$ gets data sampled, its output is mixed with data sampled from models $1\dots n$, and then the mix gets sampled to fit the model $n+1$. The uncertainty arising from all of the different modalities appearing in data causes the distribution parameters to jump around quite significantly.}%
    \label{fig:example2}%

    \centering
    \subfloat[\centering Mean estimation]{{\includegraphics[width=0.43\textwidth]{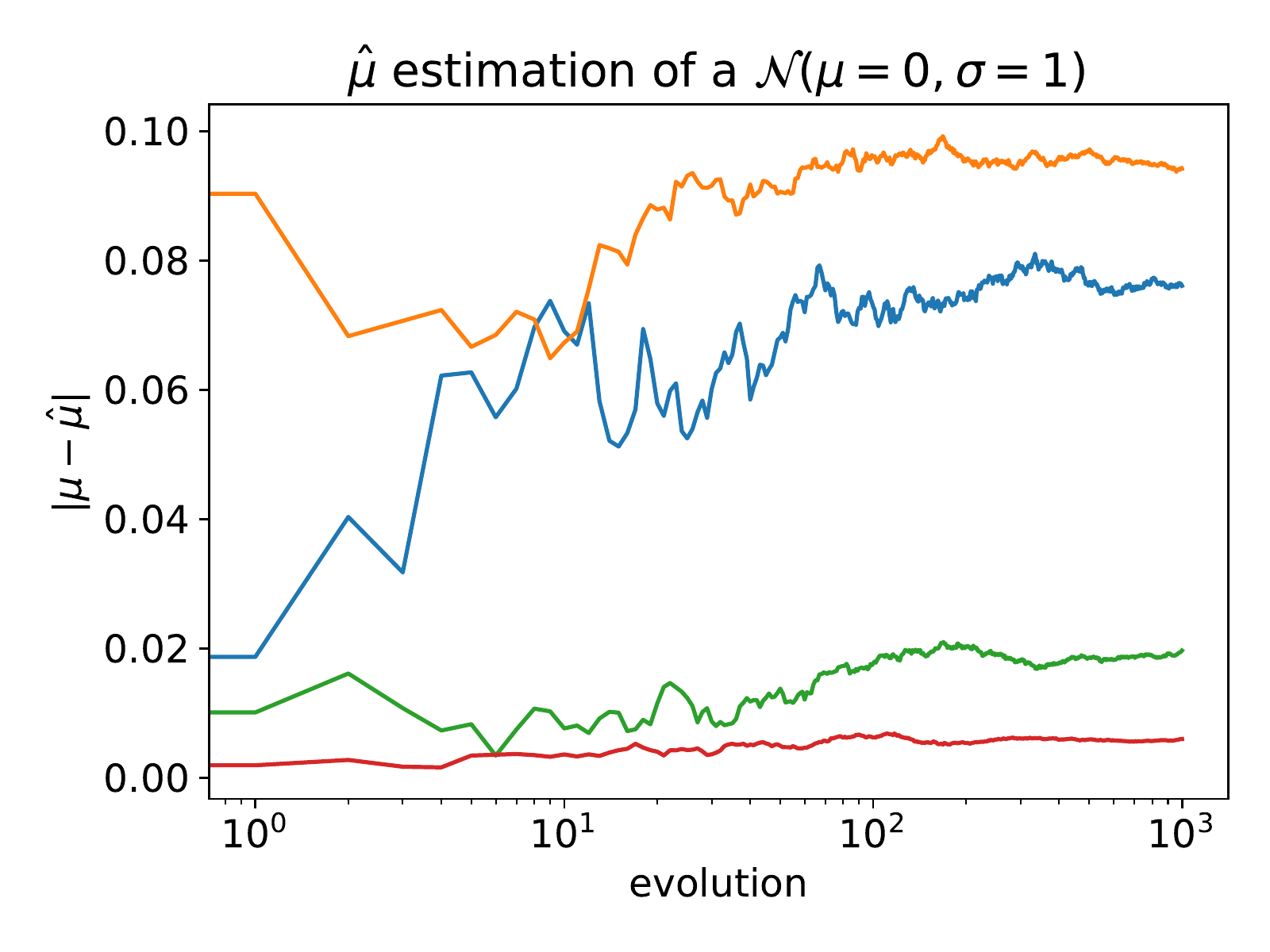} }}%
    \subfloat[\centering Standard Deviation]{{\includegraphics[width=0.43\textwidth]{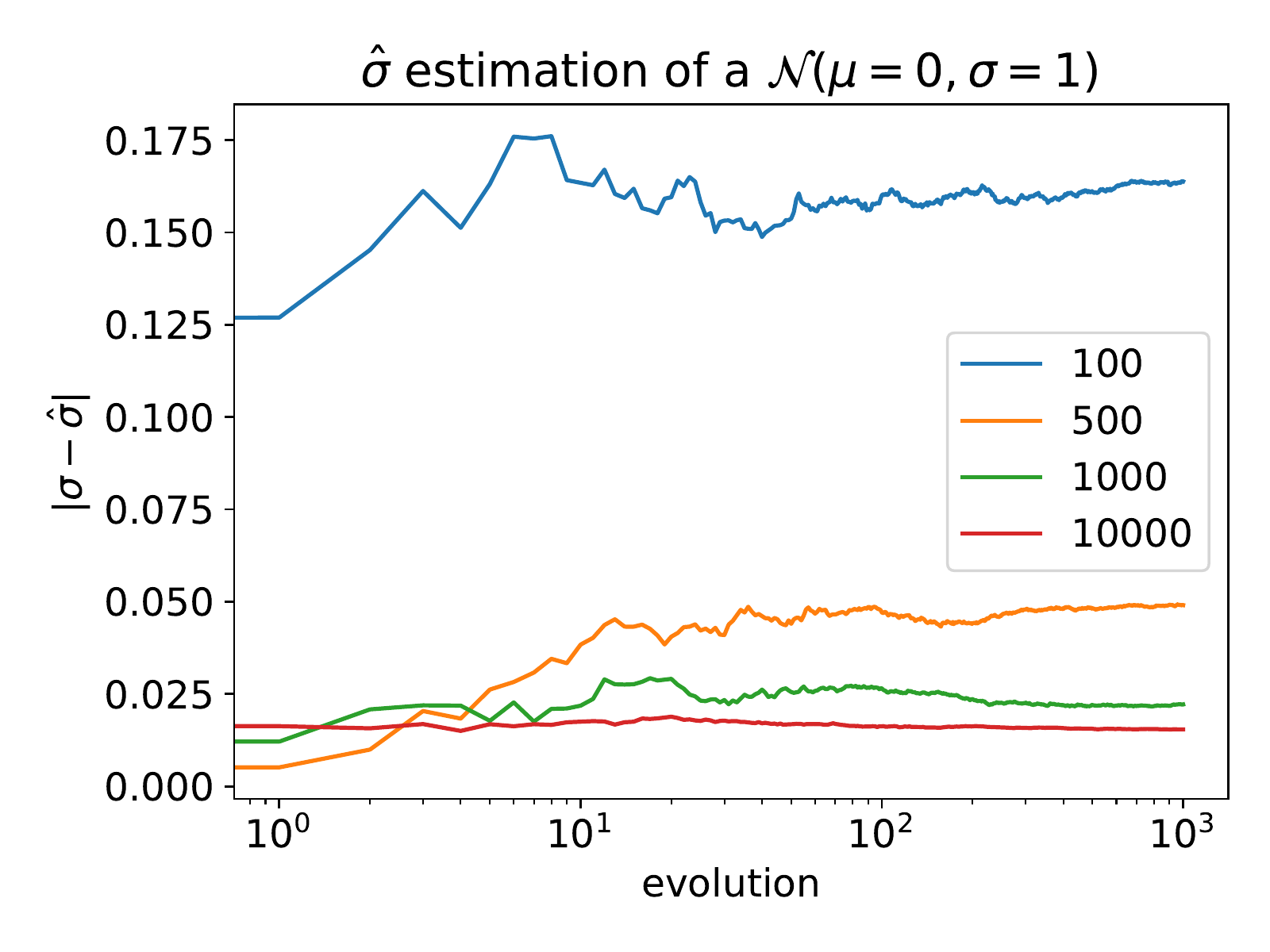} }}%
    \caption{Recursive fitting-sampling of a 1D Gaussian with different number of samples drawn. In this plot data are accumulated in a pool, all of which is used to fit a model. In other words, a model $n$ gets data sampled, its output mixed with data sampled from models $1\dots n$, and then the result is used to fit the model $n+1$. Over time the variance in estimates reduces due to linear growth of data.}%
    \label{fig:example3}%
\end{figure}

\subsection{Single dimensional Gaussian}

Following the discussion about discrete distributions, we now move on to considering how both functional approximation error and sampling error can compound (or cancel out) the process of~\md. 

To demonstrate this, consider a single dimensional Gaussian $X^0 \sim \mathcal{N}(\mu,\sigma^2)$. If we have full faith in the data we observe, the functional approximation involves estimating sample mean and variance and fitting a single dimensional Gaussian. We can estimate them using the unbiased sample mean and variance estimators:
\begin{align}
\label{eq:estimates_gaussian}
    \mu_{i+1} &= \frac{1}{M_i}\sum_j X^i_j; \quad \sigma_{i+1}^2 = \frac{1}{M_i-1}\sum _j(X^i_j-\mu_{i+1})^2.
\end{align}

Note here, that if we were to use maximum likelihood estimation, we would instead arrive at a biased variance estimator. With these estimates, the functional approximation step simply corresponds to considering a normal distribution with these parameters, which we can sample from: 
\begin{equation}
X^{i+1}_j|\mu_{i+1},\;\sigma_{i+1}\sim \mathcal{N}(\mu_{i+1},\sigma_{i+1}^2).
\end{equation}
This provides us with the conditional distribution of $X^i_j$, which allows us to calculate the full distribution of $X^i_j$. From \Cref{eq:stochproc}, we see that even after the first approximation, the distribution of $X^i_j$ is no longer normal, it follows a variance-gamma distribution \citep{fischer2023variancegamma}. However, instead of writing the probability density function at each generation, we can explicitly construct them in terms of independent random variables. In particular, it is well known \citep{cochran_1934} that $\mu_1$ and $\sigma_1$ are independent, with $\mu_1 \sim \mathcal{N}(\mu, \frac{\sigma^2}{M_0})$ and $(M_0-1)\sigma_1^2 \sim \sigma^2\Gamma(\frac{M_0-1}{2}, \frac12)$. In what follows we will denote with $Z$ random variables that are distributed with $\mathcal{N}(0, 1)$ and with $S^i$ random variables that are distributed with $\frac{1}{M_{i-1}-1}\Gamma(\frac{M_{i-1}-1}{2}, \frac12)$. 

\begin{align}
\label{eq:stochproc}
    X^0_j &= \mu + \sigma Z^0_j; \quad X^1_j = \mu + \frac{\sigma}{\sqrt{M_0}}Z^1 + \sigma\sqrt{S^1}Z^1_j; \quad \dots\\
    \nonumber X^n_j &= \mu + \frac{\sigma}{\sqrt{M_0}}Z^1 + \frac{\sigma}{\sqrt{M_1}}\sqrt{S^1}Z^2 + \dots + \frac{\sigma}{\sqrt{M_{n-1}}}\sqrt{S^1\times\dots\times S^{n-1}}Z^n+\sigma\sqrt{S^1\times\dots\times S^{n}}Z^n_j.
\end{align}

These are not joint distributions, as $Z^n$ and $S^n$ depend directly on $Z^{n-1}_j$, but when considering $X^n_j$ on its own the formula above provides all the information about the full distribution. %

The first thing we may try calculating is the variance. It is possible to find its exact value, but the mean and variance of the square root of gamma distribution are expressed in terms of gamma functions, making the result quite clunky. In what follows, we will expand everything to second order in each of $(1/M_i)$ as we assume each sample size to be large (in practice this becomes quite accurate after $M\sim100$). We then find that 
$$
\frac{1}{\sigma^2}\operatorname{Var}(X^n_j) = \frac{1}{M_0}+\frac{1}{M_1}+ \dots + \frac{1}{M_{n-1}}+1 + \mathcal{O}(2).
$$

And if we were to assume that $M_i = M$ are constant, we would find that:
$$
\operatorname{Var}(X^n_j) = \sigma^2\left(1+\frac{n}{M}\right); \quad \mathbb{E}(X^n_j) = \mu.
$$
This means that as $n\to\infty$, the variance diverges linearly. This is the same scaling as for a single dimensional Gaussian random walk. We can further see the similarities in numerical experiments shown on \Cref{fig:example1} for a range of different sample sizes, confirming these theoretical intuitions.

Even though the variance of $X^n_j$ diverges, it does not provide us with any information of what the corresponding estimates of $\mu_{n+1}$ and $\sigma^2_{n+1}$ are, or how far they are from the original $\mu$ and $\sigma$. In particular, we may want to consider what the distance would be between the true distribution and the approximated distribution at step $n+1$. To measure this we can consider the Wasserstein-2 distance between two normals:
\begin{equation*}
R^{n+1}_{W_2} := W^2_2\left(\mathcal{N}(\mu,\sigma^2),\mathcal{N}(\mu_{n+1},\sigma^2_{n+1})\right)=\|\mu_{n+1}-\mu\|^2 + \|\sigma_{n+1}-\sigma\|^2
\end{equation*}
Now we can calculate the risk that occurs due to finite sampling, \ie~what the expected value of the distance is (expanding in $1/M_i$): 
\begin{align}
\label{eq:mean_risk}
\mathbb{E}_{\mu_{n+1},\sigma_{n+1}^2}\left[R^{n+1}_{W_2}\right]&=\frac{3}{2}\sigma^2\left(\frac{1}{M_0}+\frac{1}{M_1}+ \dots + \frac{1}{M_{n}}\right)+\mathcal{O}(2),\\
\operatorname{Var}_{\mu_{n+1},\sigma_{n+1}^2}\left[R^{n+1}_{W_2}\right]&=\frac{1}{2}\sigma^4\left(\frac{3}{M_0^2}+\frac{3}{M_1^2}+ \dots + \frac{3}{M_{n}^2} + \sum_{i\neq j}\frac{4}{M_iM_j}\right)+\mathcal{O}(3).
\end{align}

This result allows us to interpret exactly what occurs in this formulation of \md. To be precise, due to errors occurring from re-sampling the approximated distribution, each generation ends up corresponding to a new step in a random walk of model parameters. The risk that occurs in this model ends up diverging for a constant sample size at each generation. In order for the end distribution approximation to be accurate, and for the distance to be finite, the sampling rate $M_i$ needs to increase superlinearly, \ie~one needs to collect increasingly more samples over time, perhaps quadratically. However, even in that case the expected distance after $n$ steps remains non-zero and the only case in which it does in fact end up being $0$ is when sampling is infinite at each step. Overall, this only shows us how far on average we go from the original distribution, but the process can only 'terminate' if the estimated variance at a certain generation becomes small enough, \ie~we effectively turn into a delta function. 

Shown on \Cref{fig:example2,fig:example3} are different runs of this process for different values of parameters of $\alpha_i, \beta_i, \gamma_i$ for different sample sizes, which was investigated numerically to see whether they can be enough to overcome \md, however even in those cases the changes are inevitable, although attenuated.

\subsection{Noisy approximation model}
\label{sec:noisyapprox}
With the simple example out of the way, we can now construct a lower bound on the distance of generation $n$ distribution from the original and show that without superlinear sampling it similarly diverges in the limit of large $n$.
A nice property of Wasserstein-2 distance is that Gaussians provide a universal lower bound for the Wasserstein distance \citep{gelbrich1990formula}. In particular, for $\kappa$ and $\nu$  probability measures on a Euclidean $N$-dimensional space with $\mu_\kappa$ and $\mu_\nu$ means, $\Sigma_\kappa$ and $\Sigma_\nu$ covariance matrices, we have that
$$
W^2_2(\kappa, \nu) \geq\left\|\mu_\kappa-\mu_\nu\right\|^2+
\operatorname{Tr}\left(\Sigma_\kappa+\Sigma_v-2\left(\Sigma_\kappa^{1 / 2} \Sigma_v \Sigma_\kappa^{1 / 2}\right)^{1 / 2}\right)\geq\left\|\mu_\kappa-\mu_\nu\right\|^2
$$
With this, instead of quantifying the distance exactly, we can instead lower bound it. The only limitation is that we are going to have to specify a functional approximation model. In order to achieve a $W_2$ bound, we will be required to specify how the mean changes between generations. In the scenario where we only have access to the sample mean, we would approximate the mean of the next generation distribution as \Cref{eq:estimates_gaussian}. However, as more information arrives, or the model begins using it better, we may end up diverging from the sample mean. We would still require that the model have good performance, \ie~on average the mean estimate is the same. We will also have to specify expected behaviour of the model over the the variance calculation, which once again will be chosen such that it averages out. Thus, we will adopt the following evolution over generations:
\begin{equation}
    \mu_{i+1}=\frac{1}{M_i}\sum_jX^i_j+\varepsilon_{i+1}=\frac{\Sigma^{1/2}_i}{\sqrt{M_i}}T^{i+1}+\mu_i+\varepsilon_{i+1}; \quad\mathbb{E}_{X^i_j}(\Sigma_{i+1})=\Sigma_i
\end{equation}
where we define $T^{i+1}$ to satisfy the equation above,~\ie~ $T^{i+1}=\frac{\Sigma^{-1/2}_i}{\sqrt{M_i}}\sum_j\left(X^i_j-\mu_i\right)$. With this normalisation $T$ has mean $0$ and covariance $I_N$ and by the central limit theorem (CLT) we would have $T^{i+1}|\mu_i,\Sigma_i\overset{\mathcal{D}}{\to}\mathcal{N}(0,I_N)$, however the lower bound will not rely on this.  
To arrive at a lower bound for the risk, similar to that of \Cref{eq:mean_risk}, we are going to have to make a few assumptions about the form of $\varepsilon_{i+1}$.\\
\textbf{Assumptions}:
\begin{enumerate}
    \item On average we can capture the mean to be the same as at the iteration prior: 
    \begin{equation}
        \mathbb{E}[\varepsilon_{i+1}|\mu_i,\Sigma_i]=0
    \end{equation}
    \item Given all of $X^i_j$, epsilon must be constant, \ie~it is a function of only the data:
    \begin{equation}
        \varepsilon_{i+1} = \varepsilon_{i+1}\left(X^i_j\right)
    \end{equation}
    In particular, it is dependent on $\mu_i$ and $\Sigma_i$ only through the data.
    \item The extra noise is orthogonal to the sample mean in the sense of random variables. This is effectively assuming that the noise does not contain any first moment information, \ie~we have:
    \begin{equation}
        \operatorname{Cov}(\varepsilon_{i+1},T^{i+1}|\mu_i,\Sigma_i)=0
    \end{equation}
    This may seem like a rather strong assumption, compared to the previous ones, however this property can be shown to hold true when imposing CLT on $T^{i+1}$ in the limit of large $M_i$ (note here that $M_i$ can only be assumed to be \textbf{large}, and not infinite) and assuming that $\varepsilon$ is strictly a function of moments higher than first. \\
    Furthermore, a property of this type is necessary to actually provide any information, since prior to it there would be no need to separate into an epsilon term and a sample mean term, since all could be absorbed into $\varepsilon$. 
\end{enumerate}
In \Cref{ap:alternative}, we further provide an alternative to Assumption 3, wherein by bounding the size of noise we are able to recover a similar bound, but only as an expansion in $1/M_i$.
\\
With all the assumptions in place, we now have the following bound:
\begin{align}
\label{eq:mean_risk_2}
\mathbb{E}\left[R^{i+1}_{W_2}\right]&\geq\mathbb{E}\left(\|\mu_{i+1}-\mu\|^2\right)\\&=\mathbb{E}\left(\|\mu_{i}-\mu\|^2\right)+\mathbb{E}\left(\|\varepsilon_{i+1}\|^2\right)+\frac{1}{M_i}\mathbb{E}\left((T^{i+1})^{\top}\Sigma_i(T^{i+1})\right)+\\&+\frac{2}{\sqrt{M_i}}\mathbb{E}\left((\varepsilon_{i+1})^{\top}\Sigma^{1/2}_iT^{i+1} + (\mu_i-\mu)^{\top}\Sigma^{1/2}_iT^{i+1} \right)\\&=\mathbb{E}\left(\|\mu_{i}-\mu\|^2\right)+\frac{\operatorname{Tr}\Sigma}{M_i}+\mathbb{E}\left(\|\varepsilon_{i+1}\|^2\right)+\frac{2}{\sqrt{M_i}}\mathbb{E}\left((\varepsilon_{i+1})^{\top}\Sigma^{1/2}_iT^{i+1} \right)
\end{align}
Now the only quantity to evaluate is 
\begin{equation}
    \frac{2}{\sqrt{M_i}}\mathbb{E}\left((\varepsilon_{i+1})^{\top}\Sigma^{1/2}_iT^{i+1}\right) 
    = \frac{2}{\sqrt{M_i}}\int d\Sigma_i\;p(\Sigma_i)\operatorname{Tr}\left[\Sigma^{1/2}_i\operatorname{Cov}(\varepsilon_{i+1},T^{i+1}|\Sigma_i)\right]=0,
\end{equation}
by Assumption 3. Therefore, the overall bound would be similar to the Gaussian case, but with extra noise variance terms:
\begin{align}
\label{eq:mean_risk_3}
\mathbb{E}_{\mu_{n+1},\sigma_{n+1}^2}\left[R^{n+1}_{W_2}\right]\geq\operatorname{Tr}\Sigma\left(\frac{1}{M_0}+\frac{1}{M_1}+ \dots + \frac{1}{M_{n}}\right)+\sum^{n+1}_{i=1} \mathbb{E}\left(\|\varepsilon_i\|^2\right)
\end{align}

As a result, we have shown that the same superlinear scaling would be required to minimise the lower bound on \md~even in the case of more generic models of approximation, in which the mean at step $i+1$ can be separated orthogonally into the sample mean and 'extra'. 

Overall, the message of this section can be summarised as follows: \\
\fbox{\begin{minipage}{\linewidth}
\emph{When learning on generational data, due to finite sampling, we are only able to \textbf{approximate} the original distribution. While on average we should recover the original distribution, the variance arising from this is non-zero. As a result, over the generations, the average distance of $n$'th generation from the original grows and can become infinite in the limit since errors compound over time.}
\end{minipage}}

\section{Evaluation}

\subsection{Training from scratch with GMMs and VAEs}
\label{sec:formscratch}

\begin{figure}
    \centering
    \includegraphics[width=\linewidth]{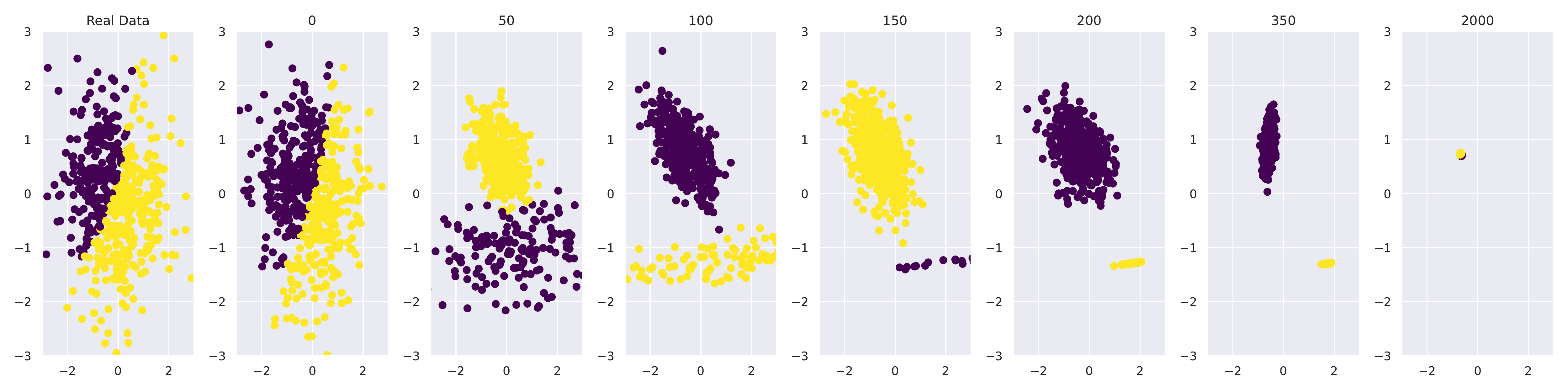}
    \caption{An examples of GMM fitting data at iterations $\{0, 50, 100, 150, 200, 350, 2000\}$. At first the model fits data very well as is shown on the left; yet even at generation 50 the perception of the underlying distribution completely changes. At generation 2000 it converges to a state with very little variance. GMM is sampled a thousand times. }
    \label{fig:gmm_evolutions}
\end{figure}

\textbf{Gaussian Mixture Models.} In this subsection we evaluate the performance of Gaussian Mixture Models (GMM) \citep{reynolds2009gaussian}. The underlying task here is that a given GMM tries to separate two artificially-generated Gaussians. \Cref{fig:gmm_evolutions} shows the progression of the GMM fitting process over time. The left-most plot shows the original two Gaussians with the ground truth labels. The next plot shows the GMM fitted on the original data with no cross-generational data used \ie~$\alpha_i = \gamma_i = 0$, where the error is minimal. Yet, within 50 iterations of re-sampling we arrive to a point where the underlying distribution is mis-perceived. The performance worsens over time and by iteration 2000 we arrive at a point estimate of the distribution with very little variance. The L2 distance between the original GMM and its descendants is plotted in~\Cref{fig:gmm_l2}.  

\begin{wrapfigure}{r}{0.5\textwidth}
  \vspace{-1em}
  \begin{center}
    \includegraphics[width=0.48\textwidth]{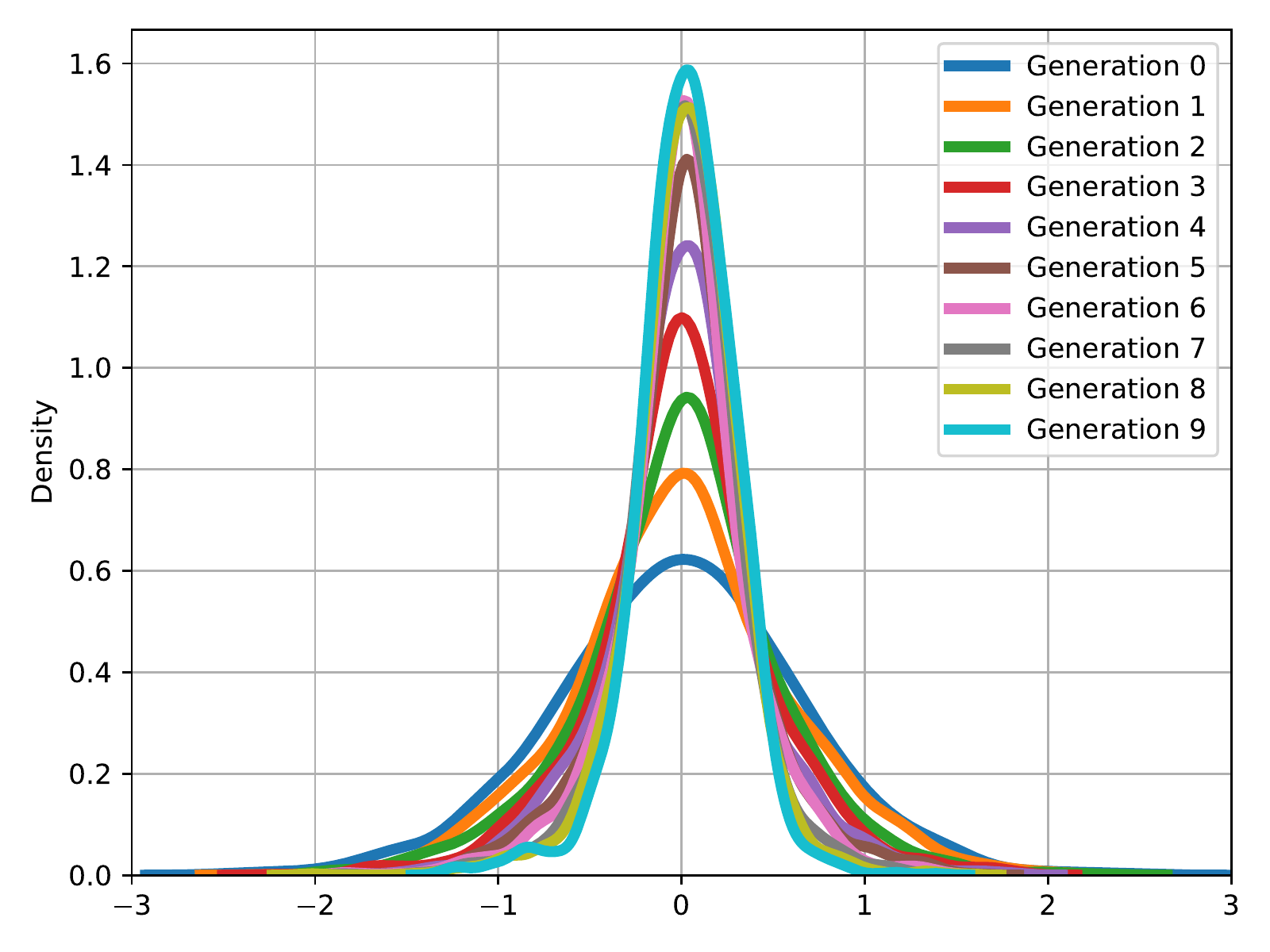}
  \end{center}
  \caption{Changing distribution of latents over the learning process with generated data as perceived by the original encoder. Just as with the Gaussian case described above, the tails get washed away and the model arrives at the mean representation of the underlying data.}
  \label{fig:density_latents}
\end{wrapfigure}

\begin{figure}%
    \centering
    \subfloat[\centering Original model]{{\includegraphics[width=0.23\textwidth]{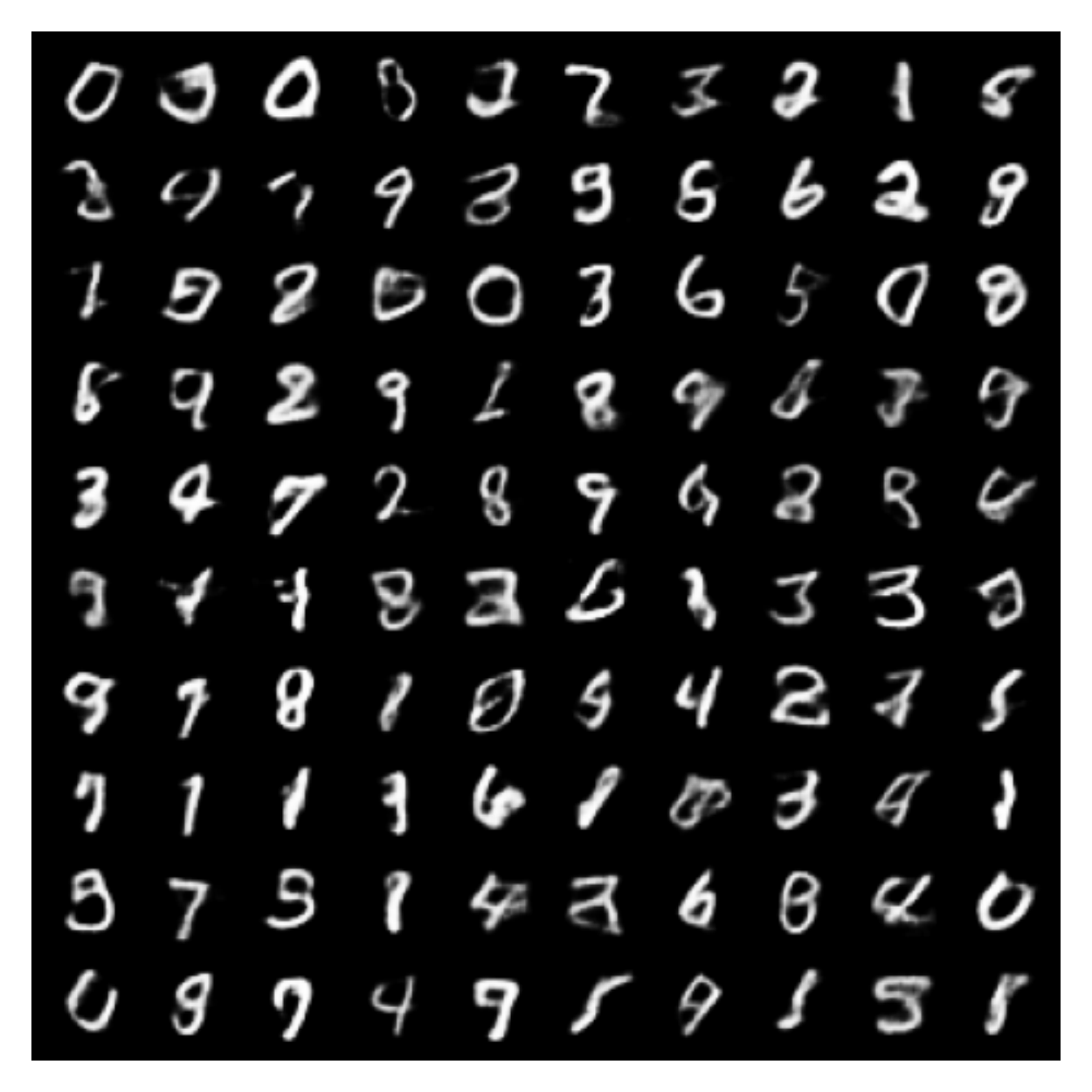} }}%
    \subfloat[\centering Generation 5]{{\includegraphics[width=0.23\textwidth]{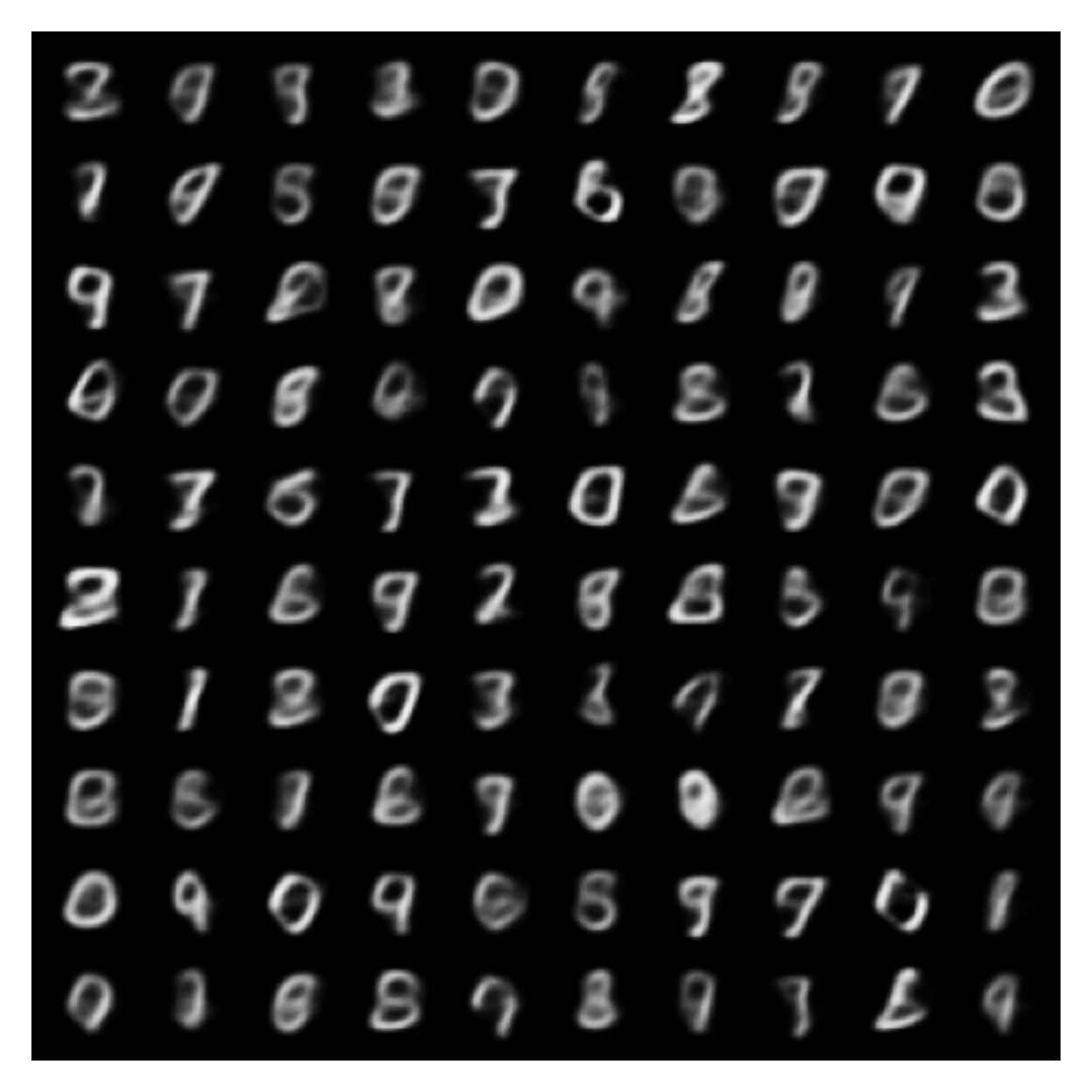} }}%
    \subfloat[\centering Generation 10]{{\includegraphics[width=0.23\textwidth]{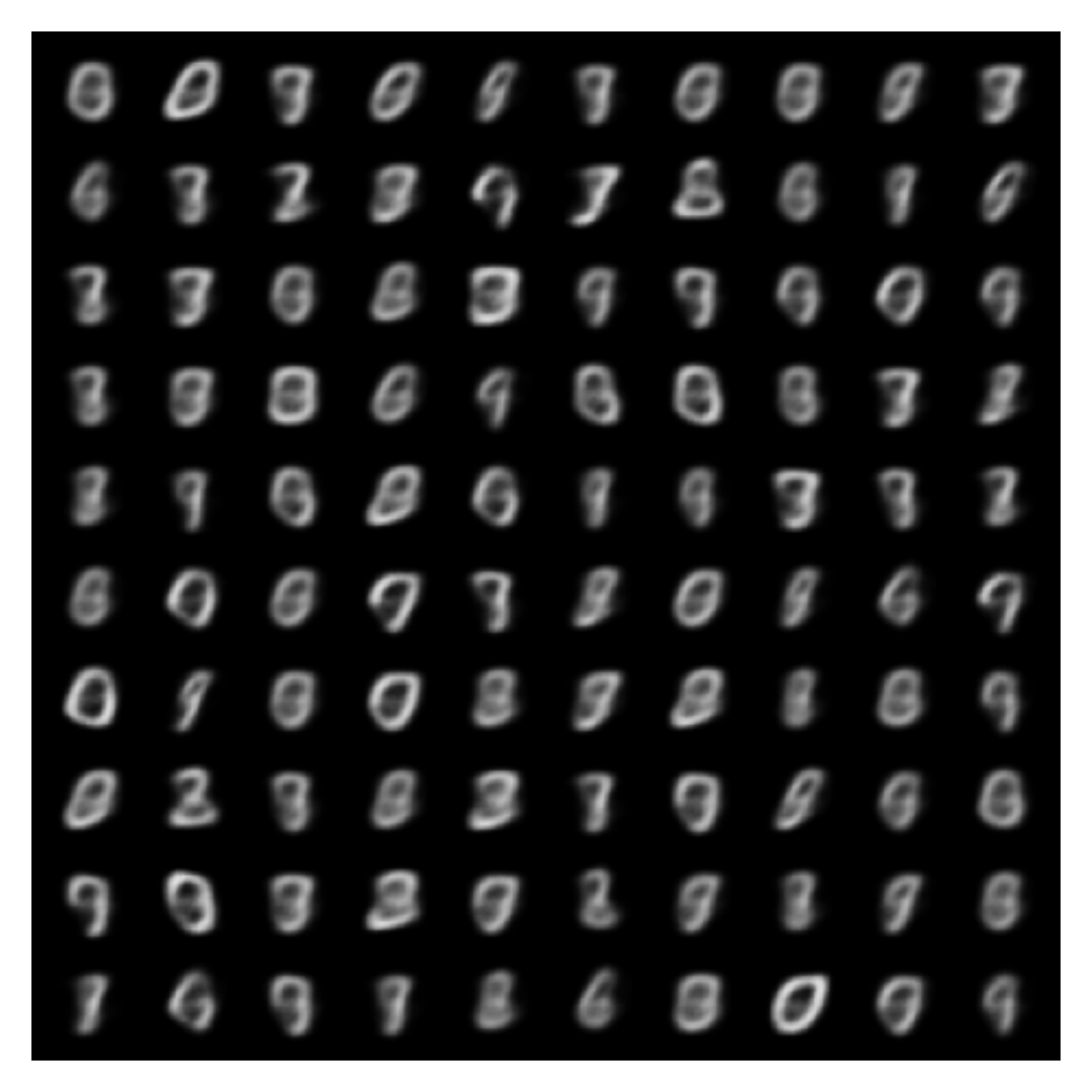} }}%
    \subfloat[\centering Generation 20]{{\includegraphics[width=0.23\textwidth]{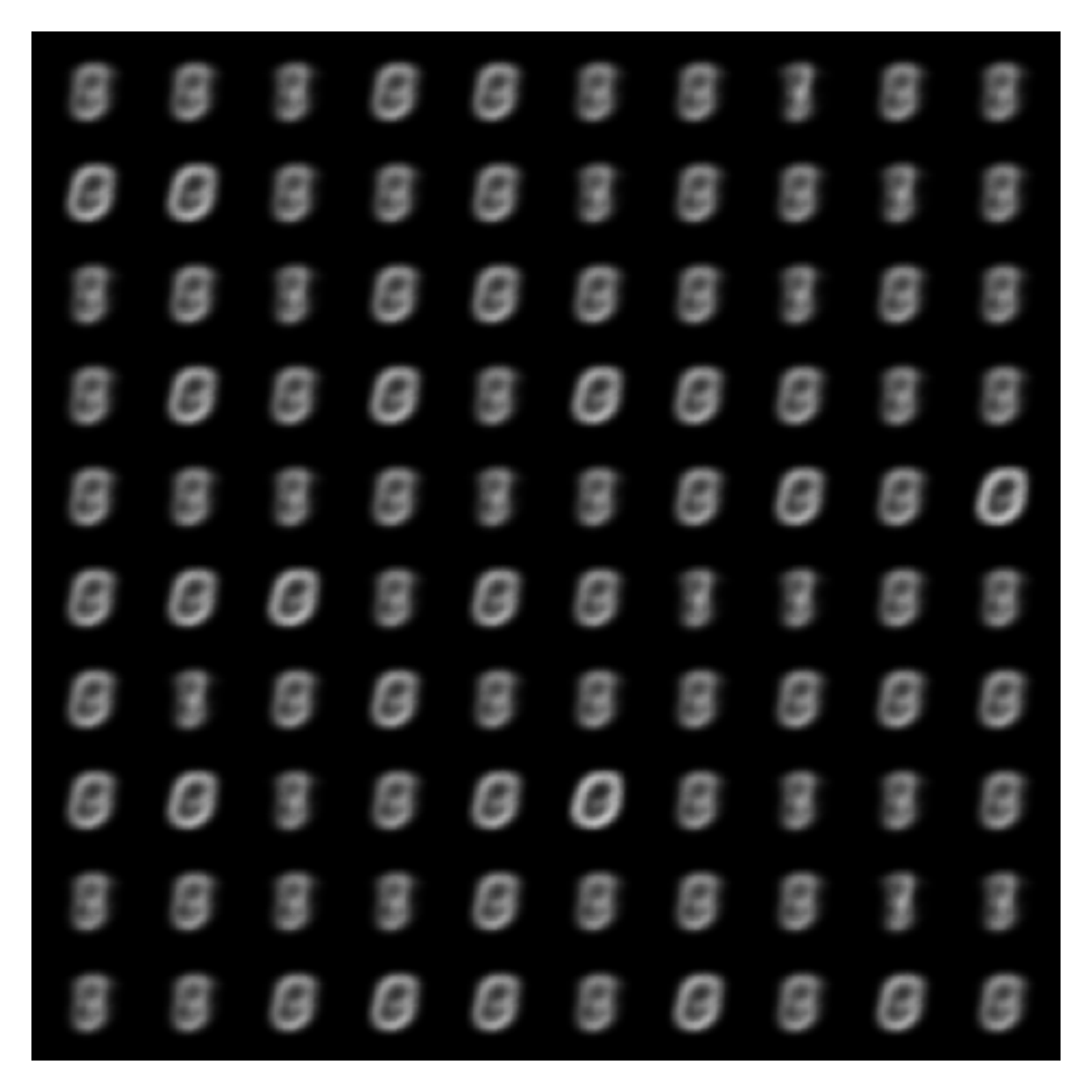} }}%

    \caption{Random latent reconstructions from VAEs. No training data comes from the original distribution. Over the generations, different modes of the original distribution get entangled and generated data starts looking unimodal.}%
    \label{fig:vae_example_latents}%
\end{figure}

\textbf{Variational Autoencoders.} In this subsection we turn to Variational Autoencoders (VAE). As before, we train an autoencoder on an original data source, which we later sample. Here, we generate latents from a Gaussian distribution which are then used by the decoder to generate data for the subsequent generation. \Cref{fig:vae_example_latents} on the left shows an example of generated data using the setting described by~\citeauthor{kingma2022autoencoding}.

Having performed the process a number of times we arrive at a representation that has very little resemblance of the original classes learned from data. On the right, one sees the generated images from generation 20, which appear to be a mix of all of the different digits. Interestingly, the original encoder perceives the generated data from its descendant with ever-growing confidence -- the encoder places such data closer and closer to the mean. \Cref{fig:density_latents} shows the density of the latent representation of the original model when presented with data generated by its descendants. As with single-dimensional Gaussians, tails disappear over time and all of the density shifts towards the mean. 

\subsection{Language Models}

By now it is clear that \Md~is universal across different families of ML models. Yet if small models such as GMMs and VAEs are normally trained from scratch, LLMs are different. They are so expensive to retrain from scratch that they are typically initialised with pre-trained models such as \texttt{BERT} \citep{devlin2018bert}, \texttt{RoBERTa} \citep{liu2019roberta}, or \texttt{GPT2} \citep{brown2020language}, which are trained on large text corpora. They are then fine-tuned to various downstream tasks~\citep{bommasani2022opportunities}. 

In this subsection we explore what happens with language models when they are sequentially fine-tuned with data generated by other models\footnote{One can easily replicate an experiment described in~\Cref{sec:formscratch} with a language model to demonstrate~\md. Given that training a single moderately large model produces twice the American lifetime worth of $CO_2$~\citep{strubell2019energy}, we opted to not run such an experiment and instead focus on a more realistic setting for a proof-of-concept. Note that just the language experiments described in the paper took weeks to run. }. We evaluate the most common setting of training a language model -- a fine-tuning setting where each of the training cycles starts from a pre-trained model with recent data. Data here comes from another fine-tuned pre-trained model. Since training is restricted to produce models that are close to the original pre-trained model and datapoints generated by the models will generally produce very small gradients, the expectation here may be that the model should only change moderately after fine-tuning. We fine-tune the \texttt{OPT-125m} causal language model made available by Meta through \texttt{Huggingface}~\citep{zhang2022opt}. 

We fine-tune the model on the \texttt{wikitext2} dataset. For data generation from the trained models we use a 5-way beam-search. We block training sequences to be 64 tokens long; then for each token sequence in the training set, we ask the model to predict the next 64 tokens. We go through all of the original training dataset and produce an artificial dataset of the same size. Since we go though all of the original dataset and predict all of the blocks, if the model had $0.0$ error it would produce the original \texttt{wikitext2} dataset. Training for each of the generations starts with generation from the original training data. Each experiment is ran 5 times and the results are shown as 5 separate runs. The original model fine-tuned with real \texttt{wikitext2} data gets $34$ mean perplexity, from the zero-shot baseline of $115$, \ie~it successfully learns the task. Finally, to be as realistic as possible, we use the best performing model on the original task, evaluated using the original \texttt{wikitext2} validation set, as the base model for the subsequent generations, meaning in practice observed~\Md~can be even more pronounced. 

\begin{figure}%
    \centering
    \subfloat[\centering No data preserved, 5 epochs]{{\includegraphics[width=0.49\textwidth]{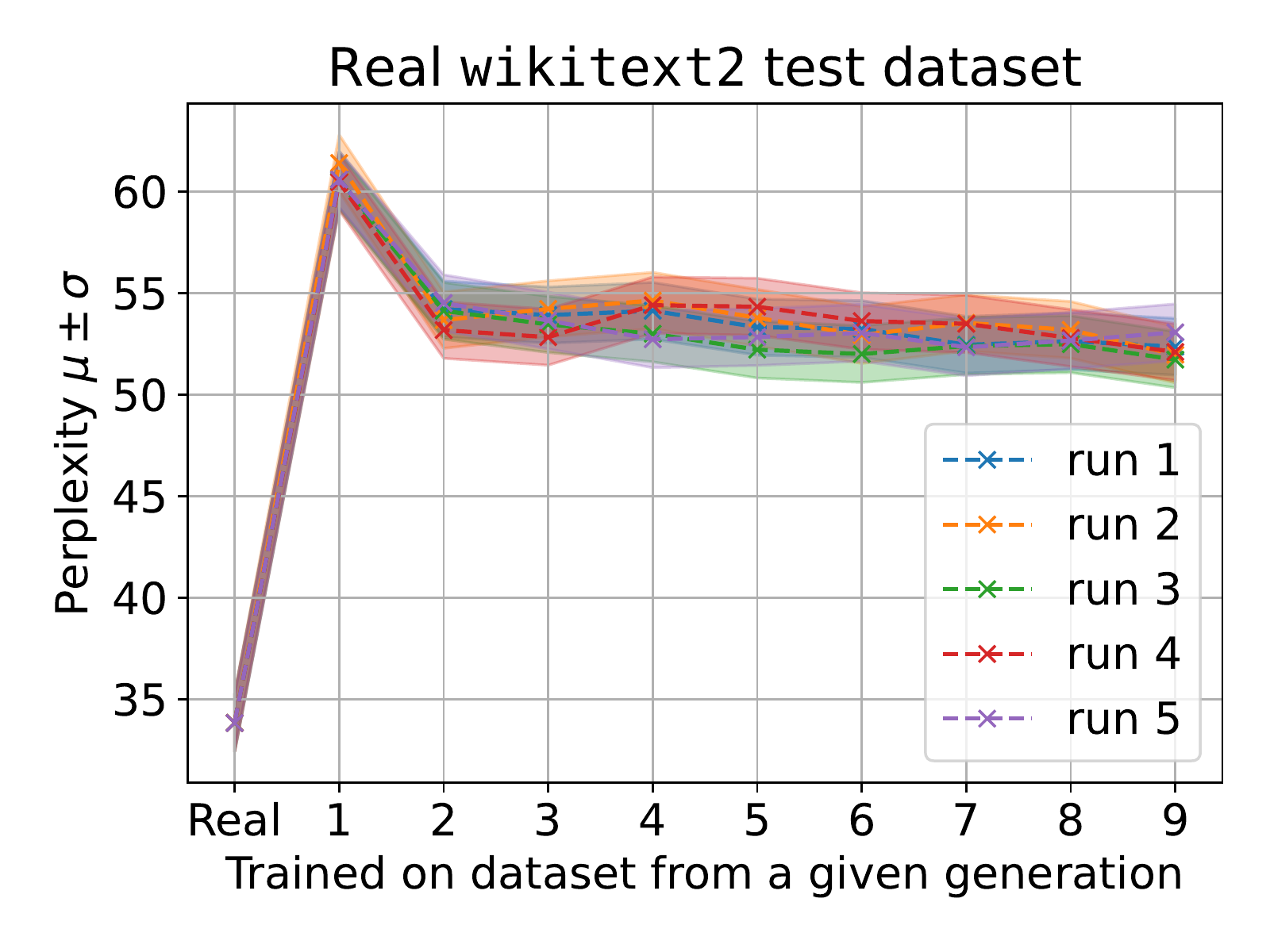} }}%
    \subfloat[\centering 10\% data preserved, 10 epochs]{{\includegraphics[width=0.49\textwidth]{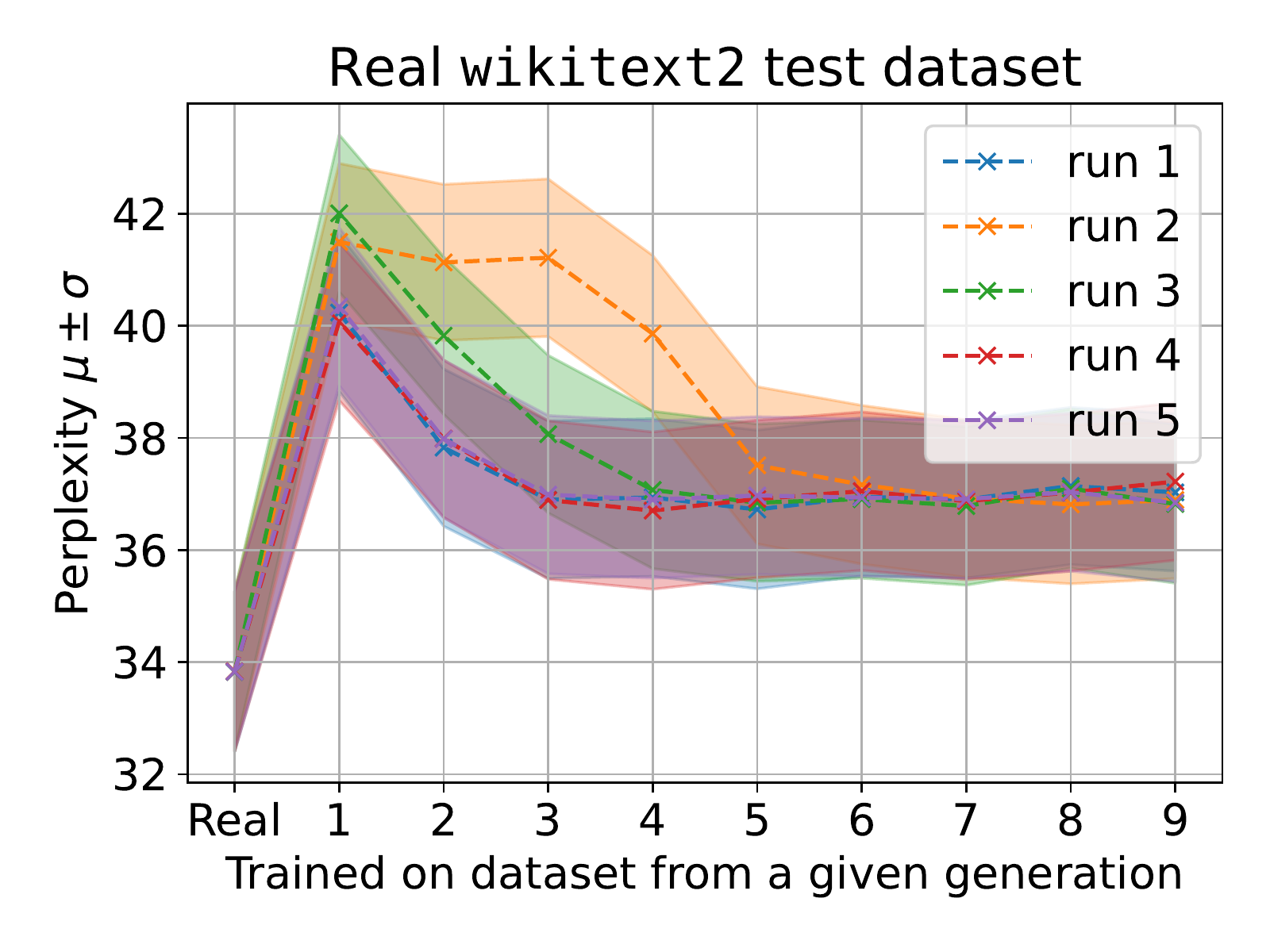} }}%
    \caption{Performance of OPT-125m models of different generations evaluated using the original \texttt{wikitext2} test dataset. Perplexity is shown on the $y$-axis and for each independent run the graph of the mean and its standard deviation is shown with error bars. $x$-axis refers to the generation of the model -- `Real' refers to the `model 0' trained on the original \texttt{wikitext2} dataset; model 1 was trained on the data produced by model 0; model 2 was trained on data produced by model 1 etc. with all generated datasets equal in size. We find that models trained on generated data are able to learn some of the original task, but with errors, as seen from the increase in perplexity. }%
    \label{fig:perp_opt125}%
\end{figure}

Here we consider two different settings: 

\textbf{5 epochs, no original training data} -- Here, the model is trained for 5 epochs on the original dataset and no original data. The overall original task performance is presented in~\Cref{fig:perp_opt125}.(a). We find that training with generated data allows one to adapt to the underlying task, losing some performance -- from $20$ to $28$ perplexity points. 

\textbf{10 epochs, 10\% of original training data preserved} -- Here the model is trained for 10 epochs on the original dataset and every new generation of training, a random 10\% of the original data points are sampled. The overall original task performance is presented in~\Cref{fig:perp_opt125}.(b). We find that preservation of the original data allows for better model fine-tuning and leads to only minor degradation of performance. 

Both training regimes lead to degraded performance in our models, yet we do find that learning with generated data is possible and models can successfully learn (some of) the underlying task. We now turn to consider  the underlying perception of probable events for each generation of our models. 

\begin{figure}%
    \centering
    \subfloat[\centering No data preserved]{{\includegraphics[width=0.49\textwidth]{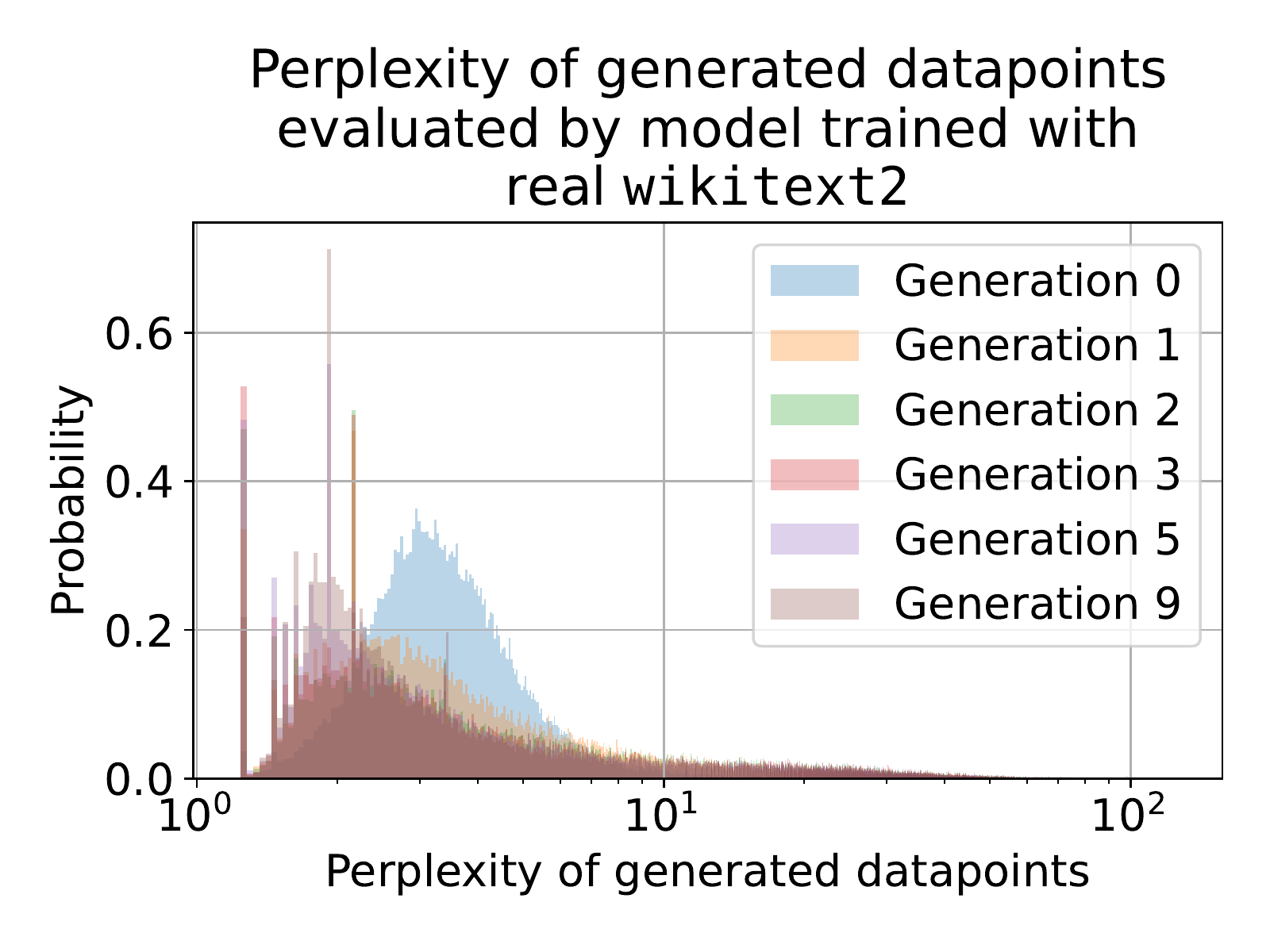} }}%
    \subfloat[\centering 10\% data preserved]{{\includegraphics[width=0.49\textwidth]{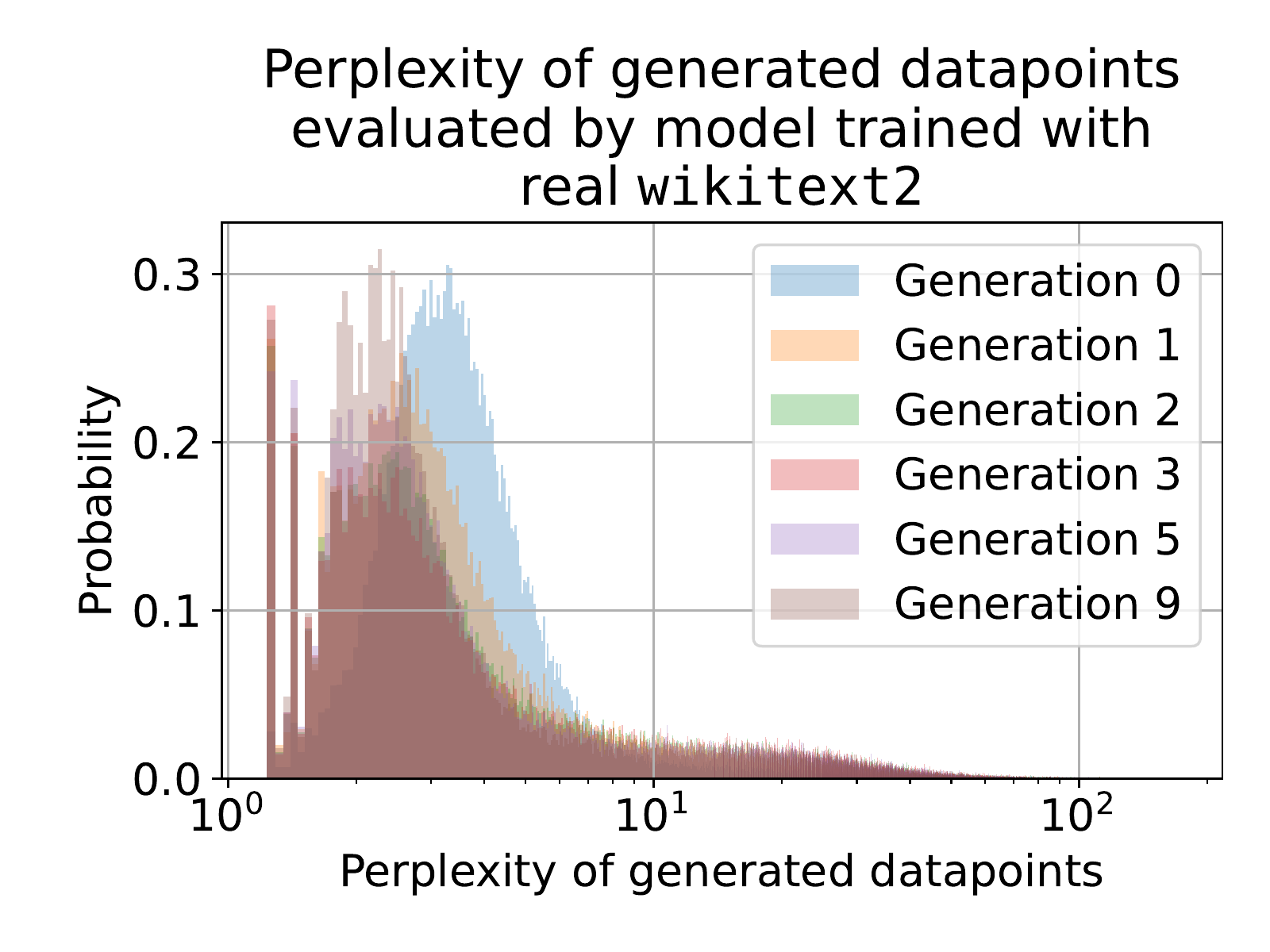} }}%
    \caption{Histograms of perplexities of each individual data training sequence produced by different generations as is evaluated by the very first model trained with the real data. Over the generations models tend to produce samples that the original model trained with real data is more likely to produce. At the same time, a much longer tail appears for later generations -- later generations start producing samples that would never be produced by the original model~\ie~they start misperceiving reality based on errors introduced by their ancestors. Same plots are shown in 3D in~\Cref{fig:pres_hist}.}%
    \label{fig:perp_hist_opt125}%
\end{figure}

\Cref{fig:perp_hist_opt125} shows histograms of individual datapoint perplexities generated by the models of different generations as is evaluated by the first model developed with real \texttt{wikitext2} training data. Here over the generations models tend to produce more sequences that the original model would produce with the higher likelihood. The observed effect is similar to that described for VAEs and GMMs in~\Cref{sec:formscratch}, where over the generations models started to produce samples that would be produced with higher probabilities by the original model. At the same time, we discover that generated data has much longer tails, suggesting that some of the data would never be produced by the original model -- these are the errors that accumulate because of the \textit{learning with generational data}. 

We find that data generated by language models in our experiments end up containing a large number of repeating phrases. The repeating problem has been observed in nearly all text generation models \citep{keskar2019ctrl,shumailov2021sponge} and to rule this out as the cause of~\Md, we further provide numerical experiments when models are explicitly encouraged to produce non-repeating sequences with repeating penalty of $2.0$. We find that this causes the models to produce lower score continuations to avoid using repeats, which as a result causes the consequent models to perform even worse. \Cref{fig:nonrepeat_hist} show model perplexities shift across the generations towards more probable token sequences. In particular, enforcing this for the LLM experiments causes the perplexity to double, compared to the original. Models remain as susceptible to~\Md, if not more.

The described process demonstrates that fine-tuning of language models does not curb the effects of~\Md~and models that are being fine-tuned are also vulnerable. We find that over the generations models tend to produce more probable sequences from the original data and start introducing their own improbable sequences~\ie~errors.

\section{Discussion and Conclusion}

We now discuss the implications of~\Md~on the underlying learning dynamics of LLMs. Long-term poisoning attacks on language models are not new. For example, we saw the creation of \textit{click, content, and troll} farms -- a form of human `language models', whose job is to misguide social networks and  search algorithms. %
The negative effect  these poisoning attacks had on search results led to changes in search algorithms: \eg, Google downgraded farmed articles\footnote{\url{https://googleblog.blogspot.com/2011/02/finding-more-high-quality-sites-in.html}}, putting more emphasis on content produced by trustworthy sources \eg~education domains, while DuckDuckGo removed them altogether\footnote{\url{https://www.technologyreview.com/2010/07/26/26327/the-search-engine-backlash-against-content-mills/}}.  

What is different with the arrival of LLMs is the scale at which such poisoning can happen once it is automated. Preserving the ability of LLMs to model low-probability events is essential to the fairness of their predictions: such events are often relevant to marginalised groups. %
Low-probability events are also vital to understand complex systems~\citep{taleb2007black}. %

Our evaluation suggests a ``first mover advantage'' when it comes to training models such as LLMs. In our work we demonstrate that training on samples from another generative model can induce a distribution shift, which over time causes~\Md. This in turn causes the model to mis-perceive the underlying learning task. To make sure that learning is sustained over a long time period, one needs to make sure that access to the original data source is preserved and that additional data not generated by LLMs remain available over time. %
The need to distinguish data generated by LLMs from other data raises questions around the provenance of content that is crawled from the Internet: it is unclear how content  generated by LLMs can be tracked at scale. 
One option is community-wide coordination to ensure that different parties involved in LLM creation and deployment share the information needed to resolve questions of provenance. 
Otherwise, it may become increasingly difficult to train newer versions of LLMs without access to data that was crawled from the Internet prior to the mass adoption of the technology, or direct access to data generated by humans at scale.

\section*{Acknowledgements}

We want to thank Anvith Thudi, David Glukhov, Peter Zaika, and Darija Barak for useful discussions and feedback.

\bibliographystyle{plainnat}  
\bibliography{bibliography}  

\newpage
\appendix
\section{Appendix}

\subsection{Absorbing Markov Chain}
\label{ap:markov}
The subsection explains a well-known fact about absorbing Markov chains, that they converge to an absorbing state with probability one. 
Assume that $\X^m$ form a Markov chain.
In order to reason about this chain we need to consider the transition probabilities. In general, these correspond to our functional approximation scheme. 
Due to the stochastic nature of the Markov chain, we expect to have the variance go up and down. But as the variance decreases, the newly sampled data, due to its finiteness, will be more concentrated, leading in the limit to a set of \ie~a delta functions. This argument assumes that the approximation scheme is good and can converge to delta functions. If not, the errors in approximation may prevent the propagation of errors in stochasticity.

As discussed in the previous section, we can model the process of repeated `sampling' and `fitting' as a Markov chain. In this subsection, we explain how such a process can converge to a stationary state \ie~the absorbing state of a Markov Chain. 
In this derivation we follow Allan Yashinski~\footnote{\url{www.math.umd.edu/~immortal/MATH401/book/ch_absorbing_markov_chains.pdf}}. Suppose we have an absorbing Markov Chain with $r$ transient states $t_1, \dots, t_r$ and $s$ absorbing states $a_1, \dots, a_s$. The whole Markov chain has $r+s$ states, ordered as follows: $t_1, \dots, t_r, a_1, \dots, a_s$. The transition matrix is then defined as 

\begin{equation}
    T = \begin{bmatrix}
            Q & 0_{r\times s}\\
            R & I_s \\
    \end{bmatrix},
\end{equation} 

where
\begin{itemize}
    \item $Q$ is an $r\times r$ matrix holds the probabilities of moving from a transient state to another transient state
    \item $R$ is an $s\times r$ matrix which holds the probabilities of moving from a transient state to an absorbing state. 
    \item $0_{r\times s}$ is the $r\times s$ matrix of all 0's. There 0's represent the probabilities of moving from an absorbing state to a transient state (which is impossible by definition).
    \item $I_s$ holds the probabilities of transitioning between the absorbing states. As transition is impossible, this is just the $s\times s$ identity matrix.
\end{itemize}

We are interested in $\lim_{k\rightarrow\infty} T^{k}(\X_0)$. For a given $k$, the matrix becomes 

\begin{equation}
    T^k = \begin{bmatrix}
            Q^k & 0_{r\times s}\\
            R + RQ + \dots + RQ^{k-1} & I_s \\
    \end{bmatrix} = \begin{bmatrix}
            Q^k & 0_{r\times s}\\
            R\sum^{k-1}_{i=0} Q^{i} & I_s \\
    \end{bmatrix}.
\end{equation} 

Finally, for an absorbing Markov chain with $T = \begin{bmatrix}
            Q & 0_{r\times s}\\
            R & I_s \\
    \end{bmatrix}$, 
    
    we have $\lim_{k\rightarrow\infty} T^{k} = \begin{bmatrix}
            0_{r\times r} & 0_{r\times s}\\
            R(I_r - Q)^{-1} & I_s \\
    \end{bmatrix}$.\\

Since in the limit the transition probabilities to transient states are zero, we end up converging to absorbing states and staying there. In the case of discrete distributions, where we can perfectly approximate a zero-variance dataset (\ie~a delta function), the absorbing states are delta functions centered at any non-zero probability point from the original distribution. In practice, we would like to know the expected number of steps before being absorbed, which may be large. But without knowing our fitting procedure it is impossible to calculate the matrix $Q$ and therefore the average length of time before collapse.

\begin{figure}
    \centering
    \includegraphics[width=\linewidth]{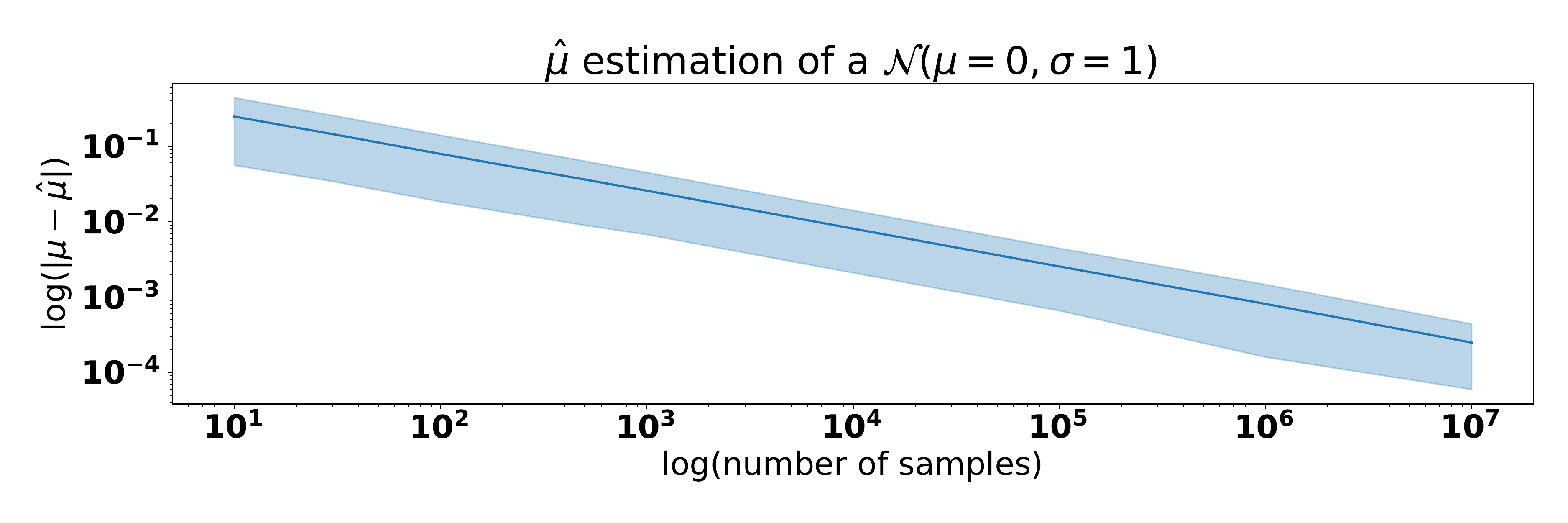}
    \caption{Approximation of a single-dimensional Gaussian $\mathcal{N}(0,1)$ as a function of number of points. The mean estimator and its standard deviation are calculated from running the procedure 10000 times.}
    \label{fig:single_dim_gaus_approx}
\end{figure}

\begin{figure}
    \centering
    \includegraphics[width=0.9\linewidth]{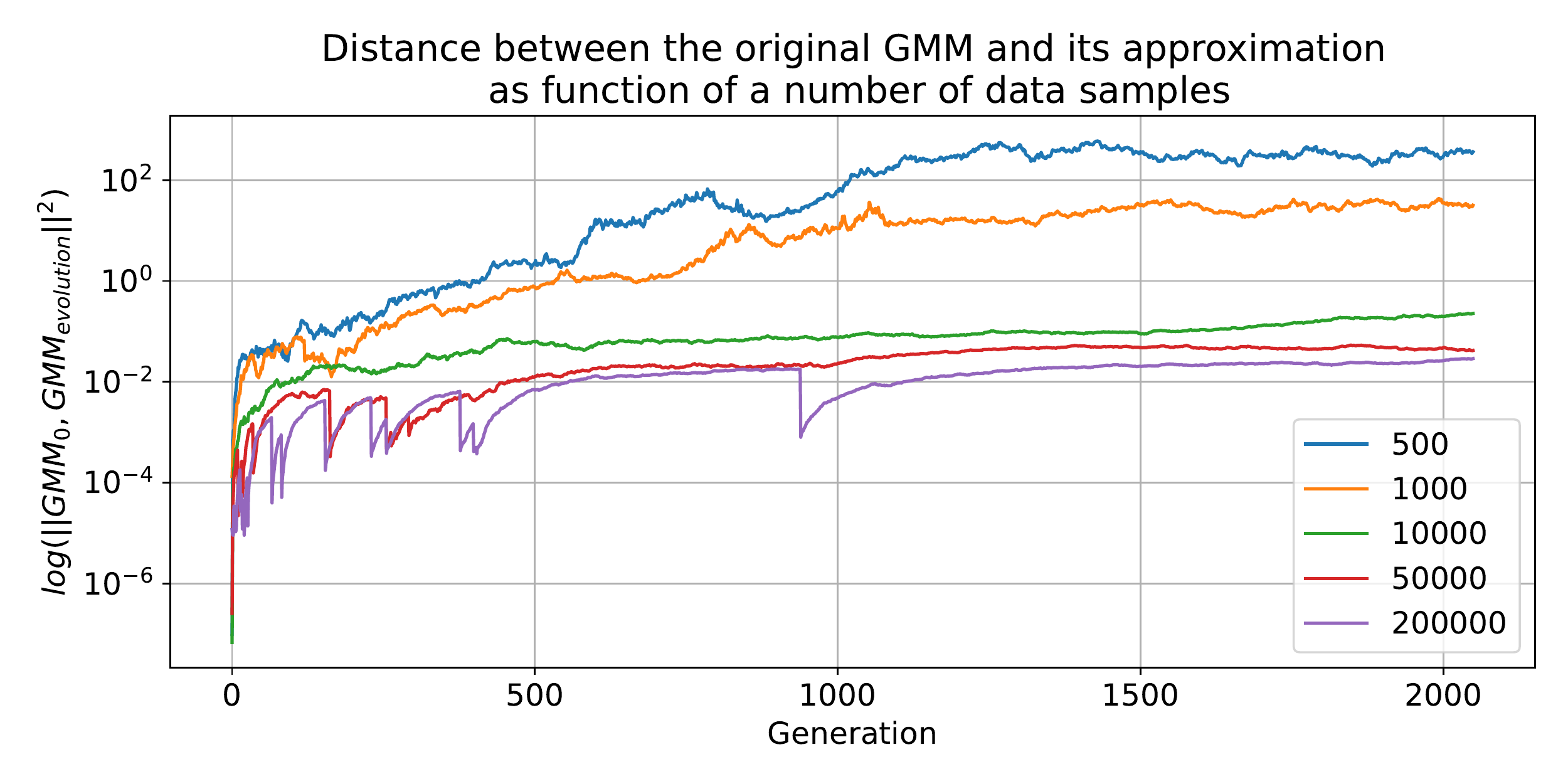}
    \caption{Progressive fitting of a GMM with different number of samples. On the $y$-axis is shown the logarithm of $L2$ distance between the two GMM distributions. Over the generations the distance begins to grow and can become quite large. The jumps in the distance for large sample sizes occur due to the fixed number of iterations and precision for the expectation maximization algorithm.}
    \label{fig:gmm_l2}
\end{figure}

\subsection{Alternative assumption for noisy approximations}
\label{ap:alternative}
This subsection will cover an alternative assumption, which may be more realistic in \textbf{some} settings, in contrast to assumption 3 from \Cref{sec:noisyapprox}, and this subsection mostly acts as an extension, rather than an alternative. In particular, instead of imposing orthogonality, we can instead impose a certain size requirement on the noise term. This in turn allows us to arrive to a similar result.

To be more precise, we will consider the same setting as in \Cref{sec:noisyapprox}, but we will now replace Assumption 3 with Assumption 3*:

\textbf{Assumptions}:
\begin{enumerate}
    \item[\textbf{3*.}] The extra noise is going to be assumed to be bounded and of the order larger than the sample mean deviation. To be precise we will have a constant $K$ (not dependent on generation $i$), such that for all $i$:
\begin{equation}
    \|\varepsilon_{i+1}\| \leq \frac{K}{M_i}
\end{equation}
\end{enumerate}
Now with the alternative assumption in place, we can follow the exact same calculations to arrive at
\begin{equation}
\label{eq:mean_risk_2_app}
\mathbb{E}\left[R^{i+1}_{W_2}\right]\geq\mathbb{E}\left(\|\mu_{i}-\mu\|^2\right)+\frac{\operatorname{Tr}\Sigma}{M_i}+\mathbb{E}\left(\|\varepsilon_{i+1}\|^2\right)+\frac{2}{\sqrt{M_i}}\mathbb{E}\left((\varepsilon_{i+1})^{\top}\Sigma^{1/2}_iT^{i+1} \right)
\end{equation}
Similar to before, we need to evaluate (which we instead bound this time):
\begin{align}
    \frac{2}{\sqrt{M_i}}\mathbb{E}\left((\varepsilon_{i+1})^{\top}\Sigma^{1/2}_iT^{i+1}\right) 
    &= \frac{2}{\sqrt{M_i}}\int d\Sigma_i\;p(\Sigma_i)\operatorname{Tr}\left[\Sigma^{1/2}_i\operatorname{Cov}(\varepsilon_{i+1},T^{i+1}|\Sigma_i)\right]\neq0\\
    &\geq -\frac{2\sqrt{N}}{\sqrt{M_i}}\int d\Sigma_i\;p(\Sigma_i)\sqrt{\operatorname{Tr}\left[\Sigma_i\Sigma_{\epsilon_{i+1}}\right]}\\
    &\geq -\frac{2\sqrt{N}}{\sqrt{M_i}}\sqrt{\mathbb{E}\left(\varepsilon_{i+1}^{\top}\Sigma_i\varepsilon_{i+1}\right)},\\
    &\geq -\frac{2\sqrt{N}}{\sqrt{M_i}}\sqrt{\frac{K^2\operatorname{Tr}\Sigma}{M_i^2}} = \frac{-2K\sqrt{N}}{M_i\sqrt{M_i}}\sqrt{\operatorname{Tr}\Sigma},
\end{align}
where we used the Cauchy-Schwarz and Jensen inequalities. Note that this is far from optimal inequality, since instead of using the expected value of the largest eigenvalue, we instead bounded it by $\operatorname{Tr}\Sigma$. 
In particular, the per step bound is then:
\begin{equation}
\label{eq:mean_risk_4_app}
\mathbb{E}\left[R^{i+1}_{W_2}\right]\geq\mathbb{E}\left(\|\mu_{i}-\mu\|^2\right)+\frac{\operatorname{Tr}\Sigma}{M_i}+\mathbb{E}\left(\|\varepsilon_{i+1}\|^2\right) - \frac{2K\sqrt{N}}{M_i\sqrt{M_i}}\sqrt{\operatorname{Tr}\Sigma}.
\end{equation}
Without knowledge of the specific values of $K$, $N$ or $\operatorname{Tr}\Sigma$, the best we can do is consider what this means for the bound as $M_i$ becomes large. In particular, contribution from the last two terms will be of order at most $3/2$. As a result we recover a bound similar to all of the ones observed so far:
\begin{align}
\label{eq:mean_risk_3_app}
\mathbb{E}_{\mu_{n+1},\sigma_{n+1}^2}\left[R_{W_2}\right]\geq\operatorname{Tr}\Sigma\left(\frac{1}{M_0}+\frac{1}{M_1}+ \dots + \frac{1}{M_{n}}\right)+\mathcal{O}(3/2)
\end{align}
In particular, we find in the same way, that superlinear scaling would be required to minimise the lower bound on \md~even in the case of more generic models of approximation, in which the mean at step $i+1$ can be separated into the sample mean and an extra bounded term of order at most $1/M_i$. 

\begin{figure}%
    \centering
    \subfloat[\centering Overlaid histograms]{{\includegraphics[width=0.49\textwidth]{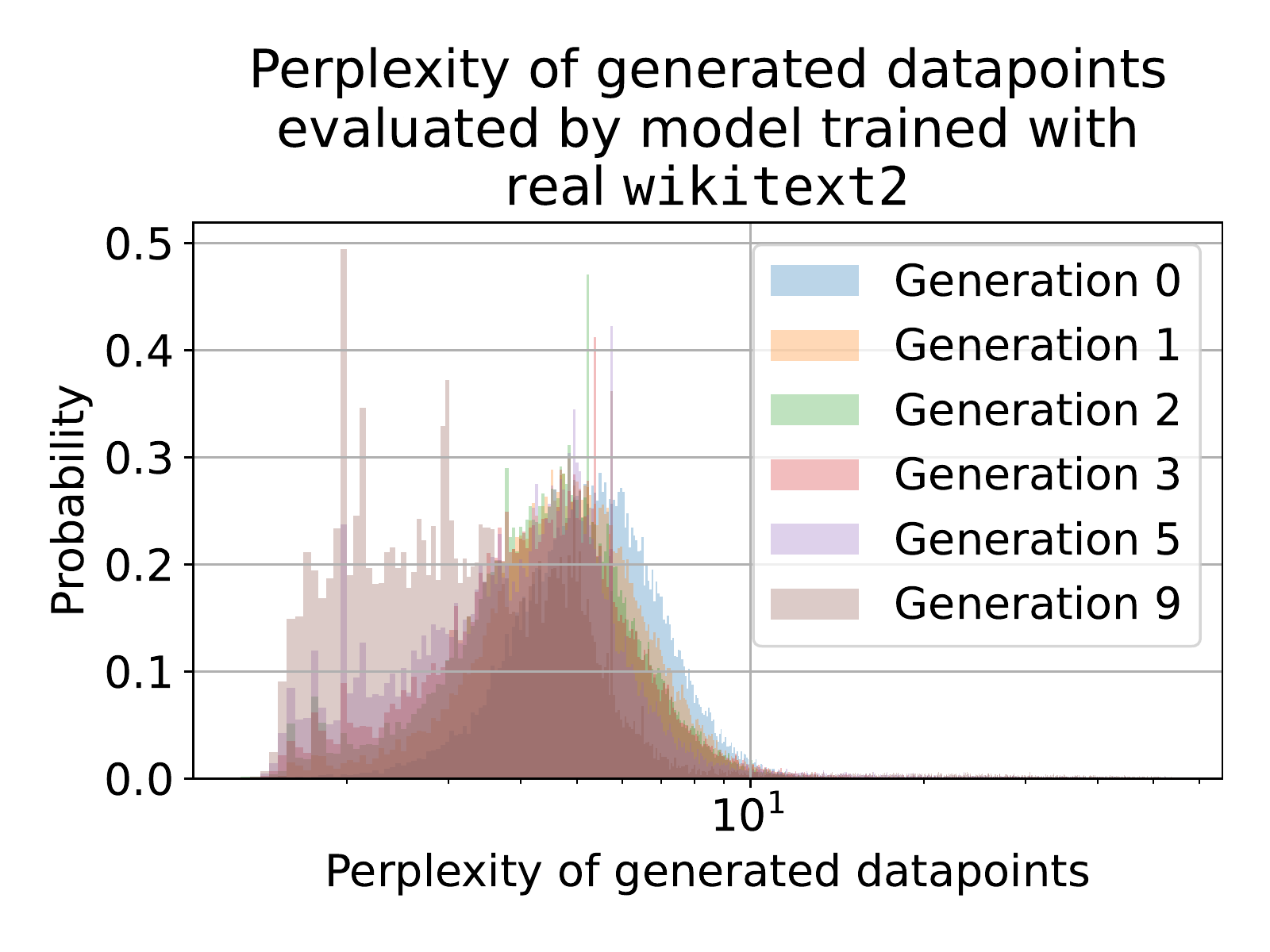} }}%
    \subfloat[\centering 3D view]{{\includegraphics[width=0.49\textwidth]{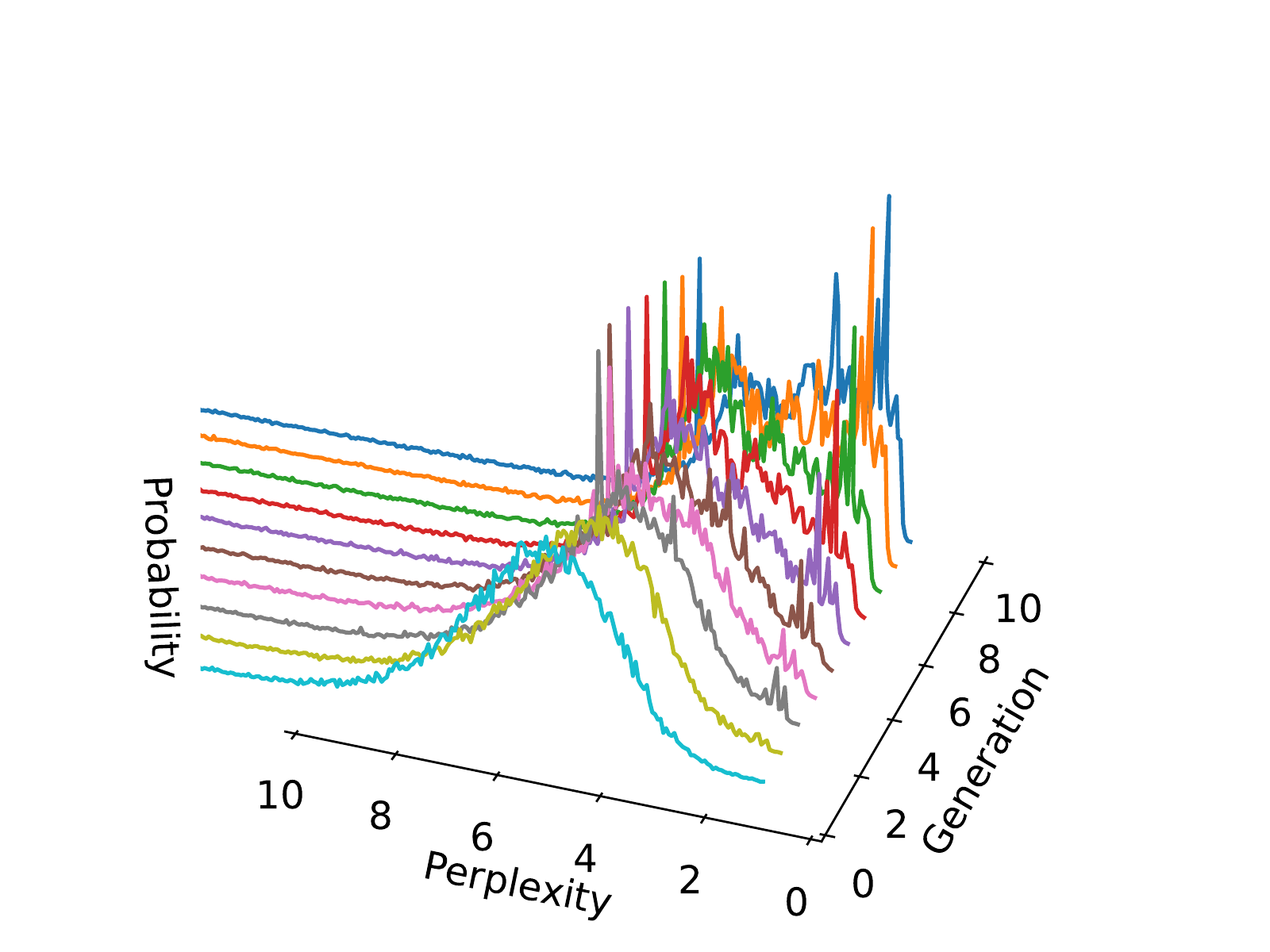} }}%
    \caption{Histogram of perplexities of each individual data training sequence produced by different generations as is evaluated by the very first model trained with the real data. Over the generations models tend to produce samples that the original model (trained with real data) is more likely to produce. At the same time, a much longer tail appears for later generations -- later generations start producing samples that would never be produced by the original model~\ie~they start misperceiving reality based on errors introduced by their ancestors. Models here are explicitly forced to not repeat sequences with a penalty of $2.0$. }%
    \label{fig:nonrepeat_hist}%
\end{figure}

\begin{figure}%
    \centering

    \subfloat[\centering \Cref{fig:perp_hist_opt125}.a in 3D. No data preserved.]{{\includegraphics[width=0.49\textwidth]{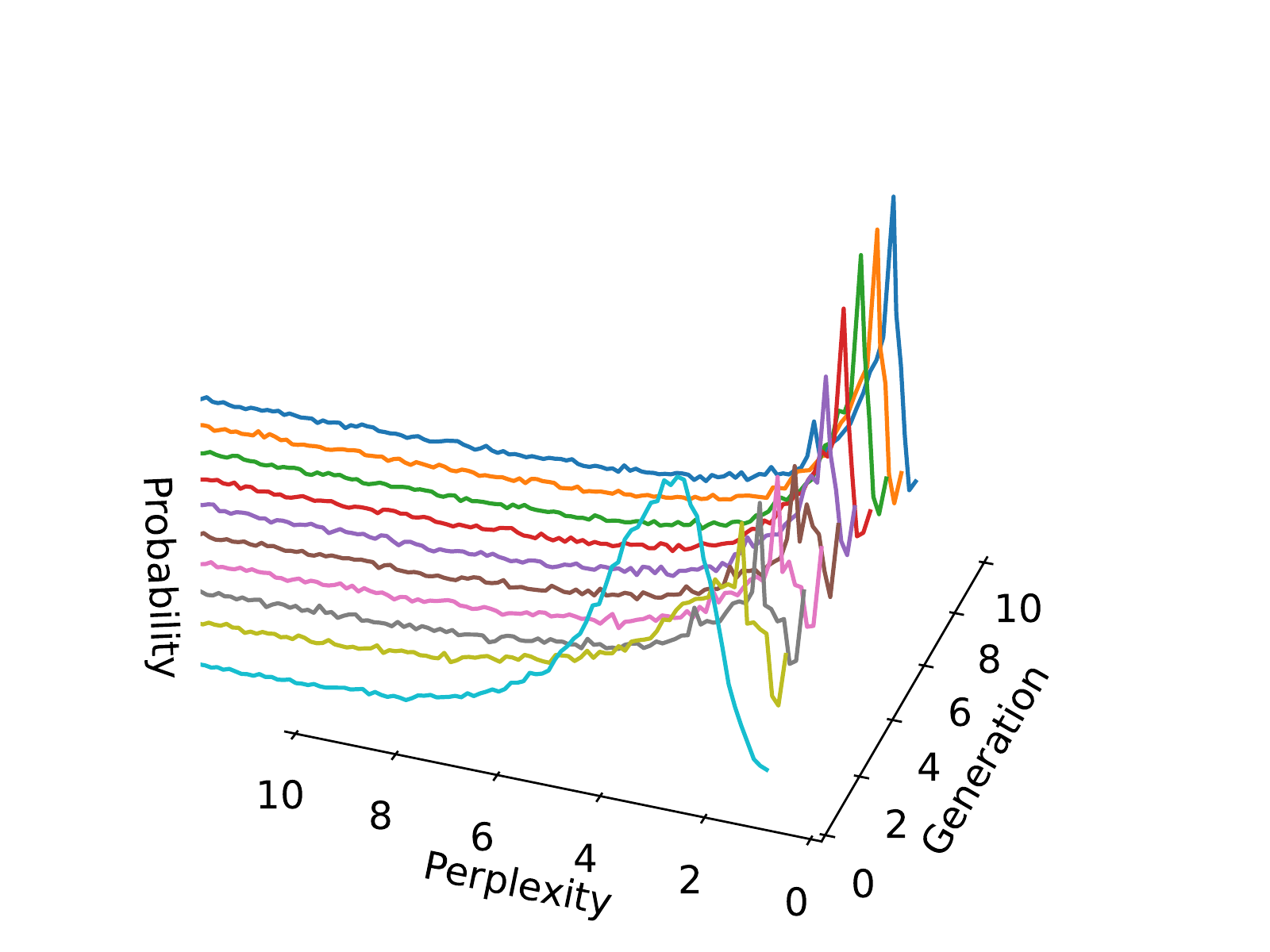} }}%
    \subfloat[\centering \Cref{fig:perp_hist_opt125}.b in 3D. 10\% original data preserved. ]{{\includegraphics[width=0.49\textwidth]{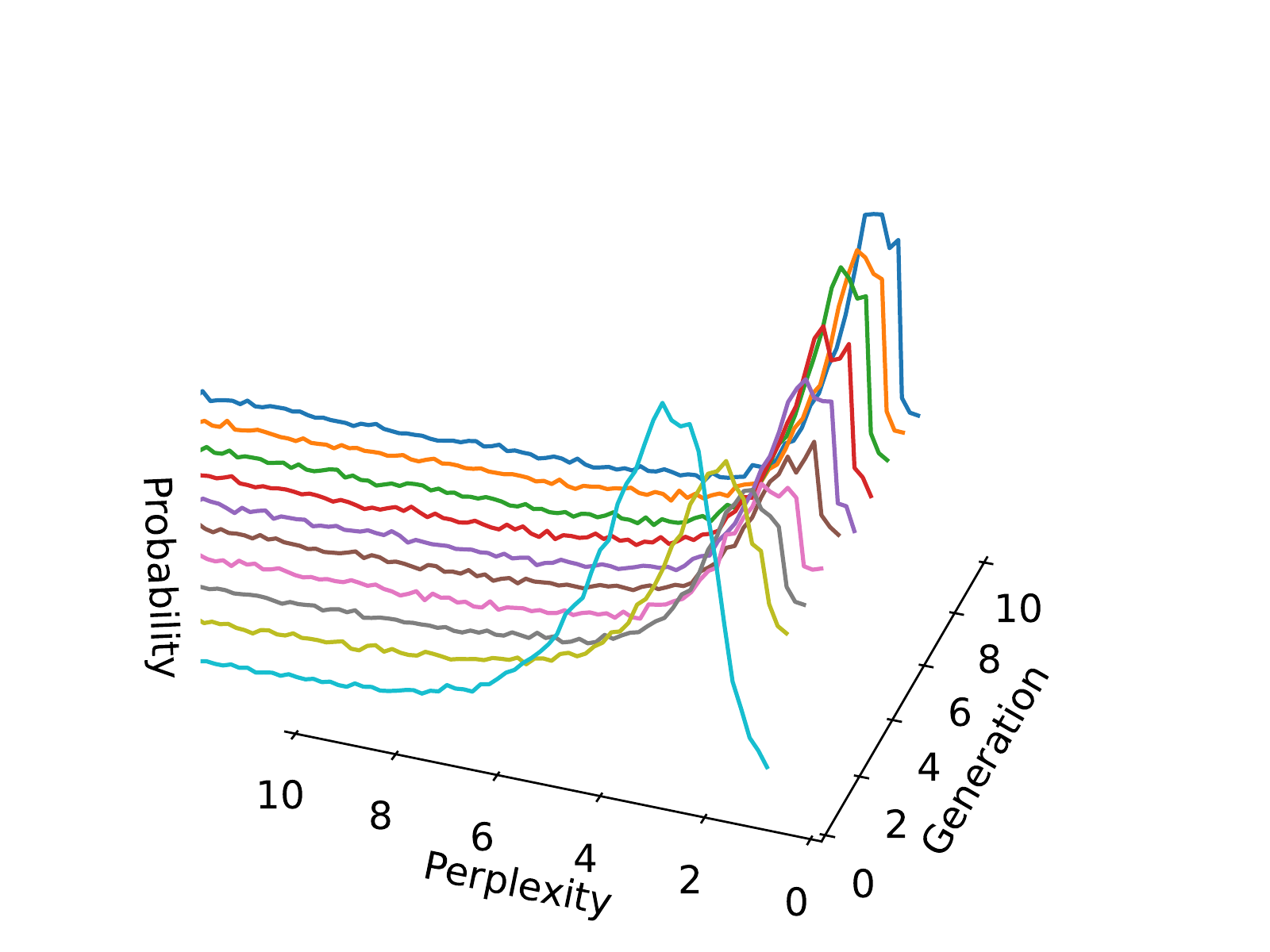} }}%
    
    \caption{Histogram of perplexities of each individual data training sequence produced by different generations as is evaluated by the very first model trained with the real data. Over the generations models tend to produce samples that the original model (trained with real data) is more likely to produce. At the same time, a much longer tail appears for later generations -- later generations start producing samples that would never be produced by the original model~\ie~they start misperceiving reality based on errors introduced by their ancestors. }%
    \label{fig:pres_hist}%
\end{figure}

\end{document}